\documentclass[journal]{IEEEtran}
\usepackage{graphicx}
\usepackage{amsmath}
\usepackage{amsthm}
\usepackage{float}
\usepackage{subfigure}
\usepackage{epstopdf}
\usepackage{cite}
\usepackage{algorithm}  
\usepackage{algorithmic}
\usepackage{multirow}
\usepackage{booktabs}
\usepackage{array}
\usepackage{bm}
\newcommand{\PreserveBackslash}[1]{\let\temp=\\#1\let\\=\temp}
\newcolumntype{C}[1]{>{\PreserveBackslash\centering}p{#1}}
\newcolumntype{R}[1]{>{\PreserveBackslash\raggedleft}p{#1}}
\newcolumntype{L}[1]{>{\PreserveBackslash\raggedright}p{#1}}

\newcommand{\tabincell}[2]{\begin{tabular}{@{}#1@{}}#2\end{tabular}}
\DeclareMathOperator*{\argmin}{\argmin} 

\makeatletter
\def\hlinew#1{%
  \noalign{\ifnum0=`}\fi\hrule \@height #1 \futurelet
   \reserved@a\@xhline}
\makeatother
\begin{document}

\title{Self-adaptive Multi-task Particle Swarm Optimization}
\author{ Xiaolong~Zheng,
			Deyun~Zhou,
			Na~Li,
	        Yu~Lei,
	        Tao~Wu,
	        Maoguo~Gong
	
	

}

\maketitle

\begin{abstract}

Multi-task optimization (MTO) studies how to simultaneously solve multiple optimization problems (i.e., tasks) for the purpose of obtaining better performance on each optimization problem. Over the past few years, evolutionary MTO (EMTO) was proposed to handle MTO problems via evolutionary algorithms. So far, many EMTO algorithms have been developed and demonstrated very well performance on solving many real-world problems. However, there remain many works to do in adapting knowledge transfer to task relatedness in EMTO. Different from the existing works, we develop a novel self-adaptive multi-task particle swarm optimization (SaMTPSO) through the developed knowledge transfer adaptation strategy, the focus search strategy and the knowledge incorporation strategy. In the knowledge transfer adaptation strategy, each task has a devised knowledge source pool that consists of all available knowledge sources. Each source (task) outputs knowledge to the task. And knowledge transfer adapts to task relatedness via individuals' choice on different sources of a pool, where the chosen probabilities for different sources are computed respectively according to task's success rate in generating improved solutions via these sources. In the focus search strategy, if there is no knowledge source benefit the optimization of a task, then all knowledge sources in the task's pool are forbidden to be utilized except the task, which helps to improve the performance of the proposed algorithm. Note that the task itself is as a knowledge source of its own. In the knowledge incorporation strategy, two different forms are developed to help the SaMTPSO explore and exploit the transferred knowledge from a chosen source, each leading to a version of the proposed SaMTPSO (i.e., SaMTPSO-S1 and SaMTPSO-S2). Several experiments are conducted on two test suites. The results of the SaMTPSO are then comparing to that of 3 other popular EMTO algorithms as well as a traditional particle swarm algorithm, which demonstrates the superiority of the SaMTPSO. 

\end{abstract}


\IEEEpeerreviewmaketitle
\section{Introduction}

\IEEEPARstart{M}{ulti-task} Optimization (MTO) \cite{GuptaOngFeng2016,zheng2019self} studies the problem of how to simultaneously solve multiple optimization problems (i.e., tasks), so as to improve on the performance of solving each task independently. It assumes that there exists some common knowledge between related tasks. This knowledge can be shared between these tasks, and help to improve the quality of new candidate solutions during the optimization process. By exploring and exploiting the relatedness between tasks, MTO makes full use of all available information that comes from one task to assist in the optimization process of another tasks, and ultimately outputs solutions with higher quality for all tasks or minimizes the time required for optimizing all these tasks. 

Studies on MTO emerged early ago. In the early years, researchers studies the possibilities of MTO using Bayesian optimization as their optimizers \cite{swersky2013multi}. Over the recent years, evolutionary MTO (EMTO) algorithms, which employs evolutionary algorithms (EAs) as optimizers, have drawn the interest of many researchers. EAs are nature-inspired search techniques \cite{585888,eiben2015evolutionary,van1998multiobjective}, which seek the optima of optimization problems by starting from a population of candidate solutions. And EAs find better solutions using various algorithmic operations during the later iterations. EMTO tries to explore and exploit the great potentials of EAs' population in benefiting the transfer of knowledge (or information) between the component tasks of MTO, so as to improve the optimization performance on these tasks. Because the useful knowledge on each task have been implicitly carried by the generated candidate solutions in population during optimization. EMTO optimizes multiple tasks simultaneously in one solver, and therefore looks a bit like the evolutionary multi-objective optimization (EMOO) which consists of multiple objects that is conflicted to each other \cite{coello2006evolutionary}. However, an EMOO problem can only be viewed as a single task that generates a set of Pareto optimal solutions. In EMTO, each task is an optimization problem, and may or may not be independent from another task. At the end of optimization, EMTO will independently output the found best solutions for each of these tasks. Therefore, EMTO is totally different from the EMOO.
 
So far, many works have been published in the field of EMTO. However, there remain many works to do in adapting knowledge transfer to task relatedness for better performance and capability. As known by many researchers of the field, knowledge transfer is very important, and should adapt to the relatedness of task pairs. Or, the negative transfer of knowledge will reduce the optimization performance on these tasks. In EMTO, inter-task knowledge transfer (knowledge are transferred across tasks) and intra-task knowledge transfer (knowledge are transferred within a task) are two very important aspects. And a EMTO solver will be able to solve optimization tasks more efficiently if intra-task knowledge transfer is assisted by a proper degree of inter-task knowledge transfer. Therefore, the existing works have studied the possibility of explicitly controlling the frequency of inter-task knowledge transfer via some parameters. By setting different parameter values to balance the inter-task knowledge transfer and intra-task knowledge transfer, no matter the manually set ones or the automatically adaptive ones, these works finally adapt knowledge transfer to the relatedness between different task pairs. Differently, we try to explore the possibility of automatically balancing these two aspects in a way that is not directly basing on the setting of parameters but basing on each task' choice on different knowledge sources (namely, component tasks).

In the proposed self-adaptive multi-task particle swarm optimization (SaMTPSO), three main strategies are developed, i.e., the knowledge transfer adaptation strategy, the focus search strategy and the knowledge incorporation strategy. At the beginning of the algorithm, the population of the SaMTPSO is split into multiple subpopulations. Each subpopulation focuses on the optimization of one unique task. Then the three main contributions, i.e., the three developed strategies, play an important role:
\begin{enumerate}
	
\item In the knowledge transfer adaptation strategy, a knowledge source pool, a success memory and a failure memory are devised for every component task respectively. A task's knowledge source pool consists of the sources outputting knowledge to the task, which are set to all available component tasks in this paper. The success memory records the number of times of each source in helping the task successfully at each generation. Similarly, the failure memory records the unsuccessful ones. With this pool and memorizes for a task, the SaMTPSO then can learn about the details of how beneficial a knowledge source is to the task by computing a success probability on the source. Based on these computed probabilities, the SaMTPSO can generate better solutions for the task by allowing the individuals of the task's subpopulation to adaptively choose different knowledge sources from the task's pool. Hence, the knowledge transfer in the SaMTPSO is self-adaptive. Besides, the individuals of the task's subpopulation will chosen different knowledge sources. Therefore, each task in the SaMTPSO can simultaneously explore and exploit the knowledge from multiple tasks (i.e., knowledge sources) at one generation. 

\item The focus search strategy handles the situation where all knowledge sources of a task fail in helping the task in generating better solutions. To implement the strategy, the SaMTPSO monitors the task's success memory. If there is no success records during the last $LP$ generations, then the strategy is activated for the task, and all knowledge sources in the task's pool are forbidden to be utilized except the task (the task is as a knowledge source of its own in the pool). That is, the strategy only allows the task's subpopulation to utilize the knowledge from the task itself. As a result, there is only intra-task knowledge transfer in this moment. Once there is any progress on the task during later optimization, the strategy will be deactivated and allow the task's subpopulation to explore and exploit all available knowledge sources within the pool. 

\item In the knowledge incorporation strategy, two independent knowledge incorporation forms are developed to help every individual of each subpopulation to make use of the transferred knowledge from its chosen source. Therefore, two versions of the SaMTPSO are developed, i.e., SaMTPSO-S1 and SaMTPSO-S2, each employing a different knowledge incorporation form. 
\end{enumerate}

The remaining paper will be organized as follows. In Section II, the related background of MTO and EMTO are first reviewed together with the particle swarm optimization. In Section III, the developed three main strategies are detailed along with the implemented SaMTPSO algorithm. In Section IV, several numerical experiments are conducted on two MTO test suites. In Section V, a brief summary for this work is presented along with some important directions for future research.

\section{Background}
\subsection{MTO and EMTO}

MTO studies the problems of how to effectively and efficiently resolve multiple optimization problems (each denoted as a component task) simultaneously. Suppose all these tasks are single-objective and unconstrained minimization problems. The optimization of an MTO problem consisting of $K$ component tasks can be mathematically presented as $\{ \bm{x}^o_1,\bm{x}^o_2,...,\bm{x}^o_K\}  = argmin\{f_1(\bm{x}_1),f_2(\bm{x}_2),...,f_k(\bm{x}_K)\}$, where the candidate solution $\bm{x}_t$ and the corresponding theoretically global optimal solution $\bm{x}^o_t$ are both feasible solutions from task $T_t$'s search space $\bm{X}_t, t=1,2,...,K$. $\bm{X}_t$ is ${D}$-dimensional. $f_t$ is the objective function of task ${T_t}$, satisfying $f_t:\bm{X}_t\rightarrow\Re$. If these tasks are constrained optimization problems, the corresponding constraint functions can be considered together with the objective functions of these tasks during optimization. Besides, if these tasks are multi-objective optimization problems, MTO will independently output the found Pareto front for each of these tasks.

%

MTO has been studied by many researchers over the years. In the very beginning, Swersky \textit{et al.} \cite{swersky2013multi} developed a multi-task Bayesian optimization framework to transfer the knowledge gained from previous optimizations tasks to new tasks, so as to find optimal hyperparameter settings for the machine learning models more efficiently \cite{bergstra2011algorithms}. In 2016, Gupta \textit{et al.} \cite{GuptaOngFeng2016} tried to transfer the biological and cultural building blocks, i.e., genes and memes, from parents to their offspring at a same time during the evolution process of population of evolutionary algorithms (EAs), and ultimately proposed a multifactorial optimization (MFO) framework as well as the implemented multifactorial evolutionary algorithm (MFEA). In MFEA, two random selected parents will mate with a probability of $rmp$, or mate if they are of the same skill factors (i.e., they are assigned to the same tasks). Therefore, each task in MFEA can explore and exploit the knowledge from all component tasks. However, knowledge transfer in MFEA can not adapt to task relatedness automatically. After the proposed MFEA, many researchers have put their interests on studying EMTO techniques. In \cite{wen2017parting}, Wen \textit{et al.} monitored the survival rate of offspring generated by two parents of different skill factors (a skill factor corresponds to a specific and unique task). Accordingly, he tried to suppress ineffective cross-task knowledge transfer when the survival rate has dropped below a manually set threshold. The implemented MFEA with resource allocation (MFEARR) has finally demonstrated promising results. In \cite{zheng2019self}, Zheng \textit{et al.} proposed a scheme to adapt knowledge transfer to the dynamically captured tasks relatedness in local region, so as to improve the performance of knowledge transfer. The scheme is then implemented as a self-regulated evolutionary MTO (SREMTO) algorithm. Besides the above works that utilized genetic algorithm (GA) as their optimizers, many researchers focused on the possibility of EMTO using particle swarm optimization (PSO). In \cite{feng2017empirical}, Feng \textit{et al.} introduced PSO into the above MFO framework to develop a multifactorial PSO (MFPSO) algorithm, which allows each particle to exploit the knowledge from other tasks with a probability of $rmp$. In \cite{RN1275}, Cheng \textit{et al.} utilized multiple populations to develop a multitasking coevolutionary PSO (MT-CPSO) algorithm, where knowledge transfer would kick in to help a particle move towards promising regions once the particle's search stagnated. Furthermore, in \cite{song2019multitasking}, Song \textit{et al.} further separated each of these multiple populations(swarms) into multiple subpopulation due to the introduction of a well-known dynamic multi-swam PSO algorithm \cite{liang2005dynamic}, and proposed a multitasking multi-swarm optimization (MTMSO) algorithm. Besides the above works, there are many other studies on methodology for EMTO \cite{da2016evolutionary,liaw2019evolutionary,2019Towards,chen2019adaptive,zhou2020toward,tang2020regularized}.


So far, many works have been done to apply the EMTO techniques into real-life application \cite{zhou2020mfea,zhao2020endmember,thanh2020efficient,liu2020multitasking}. For example, in \cite{yuan2016evolutionary}, Yuan \textit{et al.} proposed an improved MFEA algorithm to handle the MTO problem that consists of the traveling salesman task, the linear ordering task, the quadratic assignment task and the job-shop scheduling task. In \cite{RN1344}, Sampath \textit{et al.} solved the MTO problem composed of optimal power flow problems with different load demands on real-world power systems. In \cite{RN1341}, Bao \textit{et al.} developed an EMTO approach to solve cloud service composition problems.

\subsection{Particle Swarm Optimization}


Particle swarm optimization (PSO), proposed by Kennedy and Eberhart in 1995 \cite{BasicPSO}, has developed to be one of the most popular parallel evolutionary computation techniques over the past few decades \cite{poli2007particle,bratton2007defining,rini2011particle}. In PSO, each individual $ind_i, i=1,2,...,N$ is represented using a position vector ${{\bf{x}}_i} = \left\{ {{x^{d}_{i}}} \right\}_{d = 1}^D$, where $N$ is the size of population and $D$ is the dimensionality of the search space of optimization problem. To move to a new position, each individual is equipped with a velocity vector ${{\bf{v}}_i} = \left\{ {{v^{d}_{i}}} \right\}_{d = 1}^D$. Every individual will move to a new position by simply adding up its velocity ${{\bf{v}}_i}$ to its current position ${{\bf{x}}_i}$ at each generation. The update of particle's velocity ${{\bf{v}}_i}$ is regarding to three components, i.e, the inertial component, the cognition component and the social component \cite{engelbrecht2007computational}, which comes from the conceptualization of the social behaviors of bird flocking or fish schooling. The inertial component describes an individual $ind_i$'s ability to keep track of its previous direction. The cognitive component indicates the tendency of $ind_i$ to move back to the best position found by itself previously, which is denoted as personal best position ${{{\bf{pbest}}^{g}_{i}} = \{ pbest^{d,g}_{i}\} _{d = 1}^D}$ at generation $g$. And the social component considers the influence from the best position found by the whole population, which is labeled as a globally best position ${{\bf{gbest}}^g} = \{ gbest^{d,g}\} _{d = 1}^D$. Formally, the updates of particle $ind_i$'s position and velocity from generation $g$ to $g+1$ can be summarized as follows:

\begin{equation} \label{Equ:UpdateVelocity}
    \begin{aligned}
	v^{d,g+1}_{i}=wv^{d,g}_{i}+c_{1}r_{1}(pbest^{d,g}_{i}-x^{d,g}_{i})\\
	+c_{2}r_{2}(gbest^{d,g}- x^{d,g}_{i}) 
	\end{aligned}
\end{equation}
\begin{equation}  \label{Equ:UpdatePosition}
x^{d,g+1}_{i}=x^{d,g}_{i}+v^{d,g+1}_{i}
\end{equation}
, where $w$ is the inertia weight parameter for the inertial component, $c_1$ and $c_2$ are the parameters for the cognition component and the social component respectively. Usually, $w$ is set to a value linearly decreasing from $0.9$ to $0.4$. $c_1$ and $c_2$ are set to a real value within the range $[0,4]$. Coefficient $r_1$ and $r_2$ are random values within the range $[0,1]$, which satisfies the condition of a uniform distribution.

\section{The Proposed Algorithm}

\subsection{The Main Strategies of the SaMTPSO}



In this part, a novel self-adaptive multi-task particle swarm (SaMTPSO) is proposed via three developed strategies for knowledge transfer, i.e., the knowledge transfer adaptation strategy, the focus search strategy and the knowledge incorporation strategy. To design the SaMTPSO, a population is first split into multiple subpopulations evenly and randomly, namely ${\{subpop_{t}\}}_{t = 1}^K$. Each $subpop_{t}$ focuses on the optimization of one component task $T_{t}$. Therefore, each subpopulation can independently maintain the knowledge or information of its corresponding task. 

%

\subsubsection{Knowledge Transfer Adaptation Strategy}

In an MTO problem, there are multiple component tasks coexist. Supposed there are $K$ tasks in total, for each of the tasks, it can explore and exploit the knowledge from all $K$ tasks, which includes the knowledge from the task itself. For the sake of convenience, in this paper, all the $K$ tasks outputting knowledge to this task are defined as the $K$ knowledge sources (labeled as $KSs$) for this task. Then, when and which source's knowledge to make use of is what matters during the optimization of this task. Therefore, a knowledge transfer adaptation strategy is developed here for the proposed SaMTPSO. 

\begin{figure}[htb!] 
	\centering
	\includegraphics[width = 0.49\textwidth]{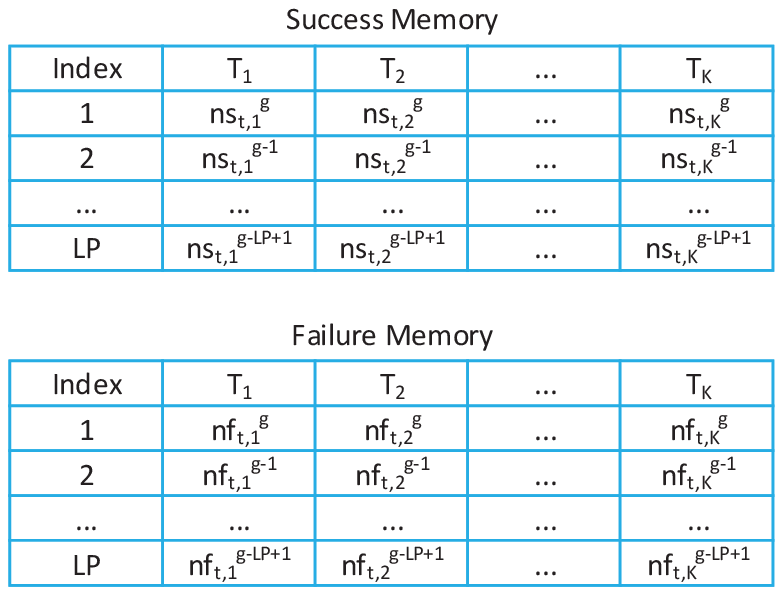}
	\caption{A component task $T_t$'s success memory and failure memory on all $K$ knowledge sources during the last $LP$ generations.}
	\label{Fig:Memories}
\end{figure}

\begin{figure}[htb!] 
	\centering
	\includegraphics[width = 0.49\textwidth]{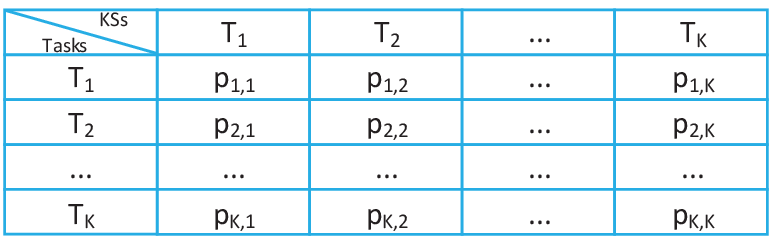}
	\caption{The learned probabilities on $K$ knowledge sources (labeled as $KSs$) for every component task at a generation $g$. }
	\label{Fig:KTR}
\end{figure}

In the strategy, the SaMTPSO tried to maintain a knowledge source pool for each of the tasks. Each pool has all the $K$ tasks as its members. In the process of optimization, every individual of a task's subpopulation will then choose one knowledge source from the task's pool according to a probability that is learned from the task's(i.e., subpopulation's) previous experience of generating promising solutions via this source. And representative information or knowledge of the chosen source will then be transferred to help the individual generating more promising solutions. More specifically, let's assume that the chosen probability of the $k$-th source in task $T_t$'s pool is $p_{t,k}, t \in [1,2,...,K], k \in [1,2,...,K]$. $p_{t,k}$ is initialized to be $1/K$, which means that all knowledge sources in $T_t$'s pool have the equal probability to be chosen at the initialization stage. Then the SaMTPSO employs the roulette wheel selection method \cite{feng2019genetic} to choose one knowledge source for every individual of $subpop_t$ respectively. The representative knowledge from these chosen sources are then utilized to generated offspring, which is discussed in the following parts. After evaluating all offspring of task $T_t$ at generation $g$, the number of offspring generated via the $k$-th knowledge source is recorded in $ns^{g}_{t,k}$ if these offspring have successfully entered the next generation. Otherwise, the number of offspring generated via the $k$-th source is recorded in $nf^{g}_{t,k}$ if these offspring are discarded in the next generation. Note that, this process is repeated for every task $T_t$. Then the SaMTPSO employs the success memories and the failure memories to store these numbers that are computed within a fixed number of previous generations(i.e., learning period ($LP$)). For example, for a task $T_t$ at the generation $g$, the number of offspring generated by different sources that can enter or fail to enter the next generation over the previous $LP$ generations are stored in different columns of the success memories and the failure memories as shown in Fig. \ref{Fig:Memories}. Once the memories overflow after $LP$ generations, the earliest records stored in the memories, i.e. $ns^{g-LP+1}_{t,k},k \in [1,2,...,K]$ and $nf^{g-LP+1}_{t,k},k \in [1,2,...,K]$, will be removed so that those numbers calculated in the current generation can be stored in these memories.

Then, the SaMTPSO update the above probabilities according to the computed success memories and the failure memories. Specifically, at the generation $g$, the probability of choosing the $k$-th source in $T_t$'s pool is updated by 
\begin{equation} \label{Equ:Norm}
{p_{t,k}} = \frac{{S{R_{t,k}}}}{{\sum\nolimits_{k = 1}^K {{SR_{t,k}}} }}
\end{equation}
, where\[{SR_{t,k}} = \frac{{\sum\nolimits_{j = g - LP+1}^{g} {{ns^{j}_{t,k}}} }}{{\sum\nolimits_{j = g - LP+1}^{g} {{ns^{j}_{t,k}}}  + \sum\nolimits_{j = g - LP+1}^{g} {{nf^{j}_{t,k}}}  + \varepsilon }} + bp\]. In ${SR_{t,k}}$, the first item represents the success rate of the offspring which are generated by the $k$-th source and have successfully entered the next generation within the previous $LP$ generations. $\epsilon=0.001$ is a small constant value used to avoid the possibility of ${\sum\nolimits_{j = g - LP+1}^{g} {{ns^{j}_{t,k}}}  + \sum\nolimits_{j = g - LP+1}^{g} {{nf^{j}_{t,k}}} } = 0 $. And the second item $bp$ is a based probability parameter to assign a small probability to those sources which haven't been chosen by any individual in $T_t$'s subpopulation within the previous $LP$ generations, such as the $k$-th source on which ${\sum\nolimits_{j = g - LP+1}^{g} {{ns^{j}_{t,k}}}  + \sum\nolimits_{j = g - LP+1}^{g} {{nf^{j}_{t,k}}} }$ equals to $0$. In equation (\ref{Equ:Norm}), the SaMTPSO divides $SR_{t,k}$ by ${\sum\nolimits_{k = 1}^K {{SR_{t,k}}} }$ to calculated ${p_{t,k}}$. So that, the sum of the probabilities of choosing different sources in every task's pool are always equaled to 1. 

From the above process of updating probability, we can know that, the larger the success rate of the $k$-th source within the previous $LP$ generations is, the larger the probability ${p_{t,k}}$ we will get. And these probabilities will be used when individuals choose knowledge sources from their corresponding pools. As a result, in every task's pool, the more successfully one source behaved in previous generations in the aspect of generating promising solutions for the task, the more probably it will be chosen in the current generation for the purpose of generating solutions. Therefore, the knowledge transfer from other tasks to a specific task is adaptive, and the frequencies of transfer between the task and other tasks (being knowledge sources in the task's pool) are decided by the number of individuals that chosen different sources respectively. Besides, the individuals of a task's subpopulation will choose different knowledge sources. Therefore, each task in the SaMTPSO can simultaneously explore and exploit the knowledge from multiple tasks at one generation. 

\subsubsection{Focus Search Strategy}









In the environment of multitasking, if a task is theoretically hard to optimize and there is nearly no relatedness between the task and other tasks, then a EMTO solver will encounter such a phenomenon: the solver is hard to find a better solution for the task with or without knowledge transfer, and knowledge transfer may probably bring negative effects. If so, the above knowledge transfer adaptation strategy will enable the SaMTPSO to explore the knowledge from all knowledge sources again and again, including the knowledge from the task itself. However, all these attempts may probably fail due to the above encountered phenomenon. Therefore, we think that,  proper times of try would be helpful in exploring potentially useful knowledge from other tasks, but it would be a waste of computing resources if the SaMTPSO keeps trying when there is no better solutions found within previous several generations. To this end, a focus search strategy is devised for the proposed SaMTPSO algorithm to monitor the occurrence of failure of knowledge transfer in every task. Once knowledge transfer with respect to all knowledge sources of a task is failed, then the focus search strategy is activated to ban all inter-task knowledge transfer for the task.

In the focus search strategy, the success memory of a component task is used to observe the occurrence of knowledge transfer failures on every knowledge source of the task. For a task $T_t$, if every $ns^{j}_{t,k} = 0$ in $T_t$'s success memory, where $k \in [1,2,...,K],j \in [1,2,...,LP]$, then the focus search strategy is activated for the task $T_t$. The activation will then allow each individual in $subpop_t$ to choose only a source ${\left\{ {{T_k}} \right\}_{k = t}}$, which means that task $T_t$ can only explore and exploit the knowledge from the task itself. Hence, there will be only intra-task knowledge transfer in this task after the activation of the focus search strategy. Note that, on the other tasks, the above knowledge transfer adaptation strategy can still work flexibly if the focus search strategy is not activated on these tasks. And the focus search strategy will then be deactivated again once any progress has been made on this task.

\subsubsection{Knowledge Incorporation Strategy}

After the above two strategies, every individual has chosen a suitable knowledge source from its corresponding task's pool. However, which kinds of knowledge or information from the source to transfer is yet to be clear. In the field of EMTO, there are multiple forms of knowledge can be transferred. A common one is the so-far found best solutions from the source, such as the globally best position ${{\bf{gbest}}^g}$ in PSO at generation $g$. The second one is to randomly choose some individuals from the sources. However, ${{\bf{gbest}}^g}$ is a more representative one as we think. In this paper, the found ${{\bf{gbest}}^g}$ for each component task $T_t$ at each generation $g$ is viewed as the representative knowledge on the task (denoted as ${{\bf{gbest}}^{g}_{t}} = \left\{ {gbest^{d,g}_{t}} \right\}_{d = 1}^{D}$), which will be transferred to other tasks once the task is chosen as a knowledge source.


To make use of the transferred knowledge from a knowledge source, a knowledge incorporation strategy is developed for the proposed SaMTPSO. Two versions of the strategy are devised as given in equation (\ref{Equ:UpdateVelocity_KT1}) or (\ref{Equ:UpdateVelocity_KT2}), which are described as:

\begin{equation} \label{Equ:UpdateVelocity_KT1}
\begin{aligned}
v^{d,g+1}_{t,i} = wv^{d,g}_{t,i} + {c_1}{r_1}(pbest^{d,g}_{t,i} - x^{d,g}_{t,i})\\
+ {c_2}{r_2}(gbest^{d,g}_{t} - x^{d,g}_{t,i})\\
+ {c_3}{r_3}(gbest^{d,g}_{ik} - x^{d,g}_{t,i})
\end{aligned}
\end{equation}

\begin{equation} \label{Equ:UpdateVelocity_KT2}
\begin{aligned}
v^{d,g+1}_{t,i}=wv^{d,g}_{t,i}+c_{1}r_{1}(pbest^{d,g}_{t,i}-x^{d,g}_{t,i})\\
+c_{2}r_{2}(gbest^{d,g}_{ik}- x^{d,g}_{t,i}) 
\end{aligned}
\end{equation}
where, $ik$ is the index of the chosen source and $t$ is the index of $subpop_t$.

Equation (\ref{Equ:UpdateVelocity_KT1}) includes four items. Compare to a traditional PSO algorithm as shown in equation (\ref{Equ:UpdateVelocity}), the forth item is to incorporate the transferred knowledge from the chosen source, i.e, $gbest^{d,g}_{ik}$ from the chosen source $T_{ik}$. However, the knowledge $gbest^{d,g}_{t}$ from a task itself has always existed as the third item of the equation (\ref{Equ:UpdateVelocity_KT1}), which will keep the influence of the task during offspring generation process. Such form of knowledge transfer has been popular in EMTO regarding to a PSO, such as the MFPSO \cite{feng2017empirical}. Therefore, equation (\ref{Equ:UpdateVelocity_KT1}) is introduced here as the first form of knowledge incorporation strategy for the proposed algorithm. 

Equation (\ref{Equ:UpdateVelocity_KT2}) only includes three items as what was in a traditional PSO algorithm. In the SaMTPSO, a component task has also been viewed as a candidate knowledge source of itself. Therefore, a second form of knowledge incorporation strategy is developed here by replacing the found $gbest^{d,g}$ of subpopulation in equation (\ref{Equ:UpdateVelocity}) with the found representative knowledge from the chosen source, i.e, $gbest^{d,g}_{ik}$ from the chosen source $T_{ik}$. According to the knowledge transfer adaptation strategy, knowledge from this chosen source has the greatest potential to improve the quantity of offspring. Therefore, this form of knowledge incorporation strategy should be efficient for the proposed algorithm.


Comparing to the second form of knowledge incorporation strategy, the first one should be less sensitive to the number of component tasks in a MTO problem due to the existing of the third item of the equation (\ref{Equ:UpdateVelocity_KT1}). However, the second one can still achieve well performance when optimizing MTO problems consisting of only a few component tasks, which has been shown in the later experiments. In this paper, we will simultaneously explore these two forms of knowledge incorporation strategies via two versions of the SaMTPSO algorithm, with SaMTPSO-S1 corresponding to the first form and SaMTPSO-S2 corresponding to the second.

\subsection{The Implementation of the SaMTPSO Algorithm}

This chapter introduces the details of the proposed SaMTPSO algorithm, which is based on a classical PSO algorithm. The structure of the SaMTPSO is shown in \textbf{Algorithm \ref{alg:Framwork}}.

\begin{algorithm}[h!]
	\caption{SaMTPSO}
	\label{alg:Framwork}
	
	\begin{algorithmic} [1]
		\REQUIRE
		~\\
		$N$ (population size)\\
		$LP$ (learning period)\\
		$bp$ (base probability parameter)\\
		$w, c_1, c_2,c_3$ (PSO parameters)\\
		\ENSURE ~\\
		$\{ {{\bf{x}}^*_1},{{\bf{x}}^*_2},...,{{\bf{x}}^*_K}\}$ (the best solutions found for the $K$ tasks)
		
		\STATE Initialize a $N$-sized population and randomly separate into $K$ subpopulations $\{ subpop_t\} _{t = 1}^K$ of size  $N_s=N/K$, including every subpopulation's positions $\{ {\bf{x}}_{t,i}\} _{i = 1}^{N_s}$ and the associated velocities $\{ {\bf{v}}_{t,i}\} _{i = 1}^{N_s}$.
		\STATE Initialize the success memory and failure memory of every task, and the corresponding probability for each knowledge source, $p_{t,k} = \frac{1}{K}, t \in [1,2,...,K], k \in [1,2,...,K]$. \label{alg:initialsrate}
		\STATE Evaluate $subpop_t$ on task $T_t$, and obtain the fitness $\{ fitness_{t,i}\} _{i = 1}^{N_s}$ and the associated personal best positions $\{ {\bf{pbest}}_{t,i}\} _{i = 1}^{N_s}$, the fitness $\{ fpbest_{t,i}\} _{i = 1}^{N_s}$, the globally best position ${{\bf{gbest}}_t}$ and the fitness $fgbest_t$.
		\STATE Generation $g=1$.
		\WHILE {(stopping conditions are not satisfied)} \label{alg:stopc}
		\STATE Generation $g \leftarrow g + 1$.
		
		\STATE \% $subpop_t$ move to new positions.
		\FOR {task $T_t, t=1$ to $K$} \label{alg:moveforward1}
		\FOR {every individual $ind_{t,i}$ of $subpop_t$}
		
		\IF {isFocus} \label{alg:firstExploration}
		\STATE Set $ ik \leftarrow t$ to perform focus search strategy. \label{alg:focusStrategy} 
		\ELSE
		\STATE Perform roulette wheel selection on $\left\{ {{p_{t,k}}} \right\}_{k = 1}^K$ to obtain the index of the chosen source $ik$. \label{alg:sklgsource}
		\ENDIF
		
		\STATE $iks_{t,i} \leftarrow ik$ to save the source for each individual.
		\STATE Incorporate knowledge by updating velocity and position (see \textbf{Algorithm (\ref{alg:knowledgeTransfer2}})).\label{alg:upositions}
		\ENDFOR
		\ENDFOR\label{alg:moveforward2}
		
		\STATE Evaluate individuals and update the corresponding success memory and failure memory (see \textbf{Algorithm (\ref{alg:Evaluations}})).\label{alg:EvaluationMemories}
		
		\STATE \% Update probability $p_{t,k}$.
		\IF {$g>LP$}
		\FOR {task $T_t, t=1$ to $K$}\label{alg:x}
		\STATE Update probability $p_{t,k}$. \label{alg:updateP} 
		\STATE Set $isFocus \leftarrow 1 $ if no success records in $T_t$'s success memory, otherwise set $isFocus \leftarrow 0 $.\label{alg:isfocus} 
		\STATE Remove $ns^{g-LP+1}_{t,k},k \in [1,2,...,K]$ and $nf^{g-LP+1}_{t,k},k \in [1,2,...,K]$ out of the memories. \label{alg:updateMs} 
		\ENDFOR
		\ENDIF
		\ENDWHILE
	\end{algorithmic}
\end{algorithm}

As previously described, the SaMTPSO splits its $N$-sized population into multiple subpopulation during initialization. The probabilities for the knowledge sources of every component task are then initialized to $1/K$, where $K$ is the number of component tasks. After initialization, the individuals in every subpopulation move to new positions. In step \ref{alg:sklgsource}, every individual from subpopulation $subpop_t$ chooses a knowledge source from task $T_t$'s pool. In step \ref{alg:focusStrategy}, only one source is allowed to be chosen for these individuals due to the activation of the focus research strategy. Then knowledge from the chosen source are employed to help the individual move to a better position in step \ref{alg:upositions}. After all individuals have moved to new positions, the proposed algorithm then evaluates the individuals of every subpopulation on the corresponding task, and update the success memories and the failure memories at the same time as shown in step \ref{alg:EvaluationMemories}. Then in step \ref{alg:updateP}, the probability for every knowledge source of a task is updated. In step \ref{alg:isfocus}, the algorithm monitors a task's success memory to find if there is any success record, and set the parameter flag $isFocus$ accordingly. Finally, the algorithm runs into another iteration if the stopping conditions are not satisfied in step \ref{alg:stopc}.

In summary, the developed knowledge transfer adaptation strategy is implemented via the step \ref{alg:sklgsource}, \ref{alg:updateP} and \ref{alg:updateMs}. The focus search strategy is implemented via the step \ref{alg:focusStrategy} and \ref{alg:isfocus}. And the knowledge incorporation strategy is via the step \ref{alg:upositions}.

\subsubsection{Updating Velocities and Positions}

In the knowledge transfer adaptation strategy and the focus search strategy, every individual has chosen a suitable knowledge source from its corresponding pool. As shown in step \ref{alg:focusStrategy} and step \ref{alg:sklgsource}, an individual of $subpop_t$ has chosen a knowledge source $T_{ik}$ from the task $T_t$'s pool. Next, the knowledge from this chosen source is used to update the individual's velocity in step \ref{alg:upositions} by the developed knowledge incorporation strategy. Note that, the found representative knowledge on task $T_t$ at each generation $g$ is denoted as ${{\bf{gbest}}^{g}_{t}} = \left\{ {gbest^{d,g}_{t}} \right\}_{d = 1}^{D}$. 
 
\begin{algorithm}[h!]
	\caption{Velocities and Positions}
	\label{alg:knowledgeTransfer2}
	
	\begin{algorithmic} [1]
		\REQUIRE
		~\\
		$ik$ (index for the chosen knowledge source)\\
		$w, c_1, c_2, c_3$ (PSO parameters)\\
		\ENSURE ~\\
		${{\bf{x}}_{t,i}}$ (individual's position)
		
		\STATE Update velocity(see \textbf{Equation (\ref{Equ:UpdateVelocity_KT1}) or (\ref{Equ:UpdateVelocity_KT2})}). \label{alg:updateVel} 
		\STATE Update position(see \textbf{Equation (\ref{Equ:UpdatePosition_KT})}).\label{alg:updatePos} 
	\end{algorithmic}
\end{algorithm}

As shown in \textbf{Algorithm \ref{alg:knowledgeTransfer2}}, individual's velocity is firstly updated in step \ref{alg:updateVel} according to equation (\ref{Equ:UpdateVelocity_KT1}) or (\ref{Equ:UpdateVelocity_KT2}), which are devised in the knowledge incorporation strategy. Then individual's position is updated according to:

\begin{equation}  \label{Equ:UpdatePosition_KT}
x^{d,g+1}_{t,i}=x^{d,g}_{t,i}+v^{d,g+1}_{t,i}  
\end{equation}
where $t$ is the index of the subpopulation $subpop_t$.

\begin{algorithm}[h!]
	\caption{Evaluations}
	\label{alg:Evaluations}
	
	\begin{algorithmic} [1]
		\REQUIRE
		~\\
		$K$ (number of tasks)\\
		$iks_{t,i}$ (the index of the chosen source by individual $ind_{t,i}$)
		\ENSURE ~\\
		$\{ \bf{x}^*_1,\bf{x}^*_2,...,\bf{x}^*_K\}$ (the best solution found on the $K$ tasks)

		\STATE \% Evaluate $subpop_t$ on task $T_t$. 
		\FOR {task $T_t, t=1$ to $K$}\label{alg:eval1}
		\FOR {every individual $ind_{t,i}$ of $subpop_t$} \label{alg2:eachIndividual}
		\STATE $fitness_{t,i} \leftarrow f_t({\bf{x}}_{t,i})$.\label{alg2:fitness}
		\STATE $ik \leftarrow iks_{t,i}$.
		
		\IF {$ fitness_{t,i} < fpbest_{t,i}$}
		\STATE ${\bf{pbest}}_{t,i} \leftarrow {\bf{x}}_{t,i}$, and $fpbest_{t,i} \leftarrow fitness_{t,i}$.\label{alg2:updatePbest}
		
		\STATE $ns^{g}_{t,ik} \leftarrow ns^{g}_{t,ik} + 1$. \label{alg2:usuccess}
		
		\IF {$ fitness_{t,i} < fgbest_{t,g}$}
		\STATE ${{\bf{gbest}}^{g}_{t}} \leftarrow {\bf{x}}_{t,i}$, and $fgbest^{g}_{t} \leftarrow fitness_{t,i}$. \label{alg2:updateGbest}
		\ENDIF
		\ELSE
		\STATE  $nf^{g}_{t,ik} \leftarrow  nf^{g}_{t,ik} + 1$. \label{alg2:ufailure}
		\ENDIF
		\ENDFOR
		\STATE Store $ns^{g}_{t,ik}$ and $nf^{g}_{t,ik}$ into the corresponding success and failure memory.
		\ENDFOR	\label{alg:eval2}
		\STATE Update the currently found best solutions $\{ \bf{x}^*_1,\bf{x}^*_2,...,\bf{x}^*_K\}$ for component tasks.
	\end{algorithmic}
\end{algorithm}

\subsubsection{Evaluating Offspring}

After updating the entire population's positions, this section evaluates all individuals on their corresponding tasks as shown in \textbf{Algorithm \ref{alg:Evaluations}}. The proposed SaMTPSO evaluates one subpopulation after another. For task $T_t$ in step \ref{alg:eval1}, the SaMTPSO evaluates the individual from the $subpop_t$ one by one in step \ref{alg2:fitness}. The $f_t$ in step \ref{alg2:fitness} is the fitness function for task $T_t$. Note that, SaMTPSO evaluates an individual on only one task. After evaluation, individual's personal best ${\bf{pbes}}{{\bf{t}}_{t,i}} = \left\{ {pbest^{d,g}_{t,i}} \right\}_{d = 1}^{D}$ and the corresponding fitness $fpbest_{t,i}$ are updated as shown in step \ref{alg2:updatePbest}. And then the $ns^{g}_{t,ik}$ is updated, if the individual has been improved by the knowledge from the chosen source $T_{ik}$. Once the $fpbest_{t,i}$ is better than the so-far found $fgbest^{g}_{t}$ on the task, then the ${{\bf{gbest}}^{g}_{t}}$ and the corresponding $fgbest^{g}_{t}$ are updated in step \ref{alg2:updateGbest}. On the contrarily, only the $nf^{g}_{t,ik}$ is updated if the individual can not be improved by the knowledge from the chosen source as shown in step \ref{alg2:ufailure}. After the individuals from a subpopulation have been evaluated, the corresponding $ns^{g}_{t,ik}$  and $nf^{g}_{t,ik}$ are stored into the corresponding success memory and the failure memory.

\section{Experiments}

In this section, four experiments will be conducted to study the performance of the proposed SaEMTO. Note that, two versions of the implemented SaMTPSO algorithms are presented, i.e., SaMTPSO-S1 and SaMTPSO-S2. The effectiveness, efficiency and adaptation property of knowledge transfer of the proposed algorithm are first studied by comparing the performance of the SaMTPSO algorithm with that of a traditional PSO algorithm \cite{BasicPSO}. The PSO algorithm has the same evolution operations and parameter settings to the SaMTPSO, except the knowledge transfer adaptation strategy, the focus search strategy and the knowledge incorporation strategy in the SaMTPSO. Then the performance of the SaMTPSO on a popular bi-task MTO test suite are presented in comparing to that of the MFPSO, the SREMTO and the MFEA. Furthermore, the performance of the SaMTPSO on handling 5-task MTO test problems are presented in comparing to that of the PSO, the SREMTO, the MFEA, which can show us the efficiency of the proposed SaMTPSO on handling MTO problems consisting of more than 2 component tasks. In the final experiment, parameter analysis are performed on the parameter $bp$ and $LP$ respectively. The impacts of these parameters are shown after comparing the performance of the SaMTPSO on the popular MTO test suite under different parameter settings. 

\subsection{Test Problems}

Two test suites will be used in the following experiments. Test suite 1 contains 9 popularly used MTO test problems. As shown in Table \ref{tab:benchmarkset}, each of these well-designed MTO problems consists of two distinct single-objective optimization component tasks. More specifically, these MTO problems various a lot in the degree of intersection and the inter-task similarity. The degree of intersection indicates the degrees of having the same dimensional element values in the two component tasks' global optima. As given in the table, these 9 MTO problems have been classified into 3 categories, i.e., complete intersection, partial intersection and no intersection. Complete intersection means that the two component tasks of an MTO problem have the same global optima. And no intersection indicates that there are no the same element values in all dimensions of these tasks' global optima. Accordingly, partial intersection means there are the same element values only in some dimensions. The inter-task similarity ($R_s$) measures the relatedness between the two component tasks of an MTO problem using the Spearman's rank correlation coefficient \cite{MFOSC2016}. $R_s$ is of range [-1, 1], and $R_S=0$ means that there is no relatedness between the two tasks, $R_s=1$ means that the two component tasks are highly related. A positive sign in $R_s$ indicates positive relatedness, and negative sign means negative relatedness. In test suite 1, there are high inter-task similarity between the tasks in problem 1, 4, and 7, while there are nearly no inter-task relatedness in problem 3, 6 and 9.

\begin{table}[htb!]
	\centering
	\caption{Description of the 9 MTO test problems in test suite 1. $D$ indicates the dimensionality of each task. $R_s$ denotes inter-task relatedness.}
	\begin{tabular}{C{0.06\textwidth} L{0.12\textwidth} C{0.12\textwidth} C{0.04\textwidth} C{0.04\textwidth}}
		
		\toprule 
		\multirow{2}{*}{ \tabincell{c}{MTO \\Problem}} &\multirow{2}{*}{\tabincell{c}{Component Task}}&\multirow{2}{*}{\tabincell{c}{Degree of intersection}}&\multirow{2}{*}{\tabincell{c}{$D$ }}&\multirow{2}{*}{\tabincell{c}{$R_s$}}\\
		&&&&\\
		
		\midrule 
		\multirow{2}[1]{*}{1} & $T_1$: Grewank & \multirow{2}[1]{*}{Complete intersection} & \multirow{2}[1]{*}{50} & \multirow{2}[1]{*}{1.00} \\
		& $T_2$: Rastrigin &       &       &  \\
		\cmidrule{1-5}    \multirow{2}[2]{*}{2} & $T_1$: Ackely & \multirow{2}[2]{*}{Complete intersection} & \multirow{2}[2]{*}{50} & \multirow{2}[2]{*}{0.23} \\
		& $T_2$: Rastrigin &       &       &  \\
		\cmidrule{1-5}    \multirow{2}[2]{*}{3} & $T_1$: Ackely & \multirow{2}[2]{*}{Complete intersection} & \multirow{2}[2]{*}{50} & \multirow{2}[2]{*}{0.00} \\
		& $T_2$: Schwefel &       &       &  \\
		\cmidrule{1-5}    \multirow{2}[2]{*}{4} & $T_1$: Rastrigin & \multirow{2}[2]{*}{Partial intersection} & \multirow{2}[2]{*}{50} & \multirow{2}[2]{*}{0.87} \\
		& $T_2$: Sphere &       &       &  \\
		\cmidrule{1-5}    \multirow{2}[2]{*}{5} & $T_1$: Ackely & \multirow{2}[2]{*}{Partial intersection} & \multirow{2}[2]{*}{50} & \multirow{2}[2]{*}{0.22} \\
		& $T_2$: Rosenbrock &       &       &  \\
		\cmidrule{1-5}    \multirow{2}[2]{*}{6} & $T_1$: Ackely & \multirow{2}[2]{*}{Partial intersection} & \multirow{2}[2]{*}{50($T_2$: 25)} & \multirow{2}[2]{*}{0.07} \\
		& $T_2$: Weierstrass &       &       &  \\
		\cmidrule{1-5}    \multirow{2}[2]{*}{7} & $T_1$: Rosenbrock & \multirow{2}[2]{*}{No intersection} & \multirow{2}[2]{*}{50} & \multirow{2}[2]{*}{0.94} \\
		& $T_2$: Rastrigin &       &       &  \\
		\cmidrule{1-5}    \multirow{2}[2]{*}{8} & $T_1$: Griewank & \multirow{2}[2]{*}{No intersection} & \multirow{2}[2]{*}{50} & \multirow{2}[2]{*}{0.37} \\
		& $T_2$: Weierstrass &       &       &  \\
		\cmidrule{1-5}    \multirow{2}[2]{*}{9} & $T_1$: Rastrigin & \multirow{2}[2]{*}{No intersection} & \multirow{2}[2]{*}{50} & \multirow{2}[2]{*}{0.00} \\
		& $T_2$: Weierstrass &       &       &  \\
		
		\bottomrule
	\end{tabular}%
	\label{tab:benchmarkset}
\end{table}

Test suite 2 consists of 9 5-task MTO problems as shown in Table \ref{tab:manytask_set}. In each of these MTO problems, the 5 component tasks are all 50-dimensional single-objective optimization problem and have no intersection on their global optima. In MTO problem 1 to 3, the 5 component tasks are constructed using a same optimization function but with different shifts and rotations. For example, the 5 component tasks in MTO problem 1 are constructed with a $Shpere$ function of different shifts and rotations. In MTO problem 4 to 7, the 5 component tasks are constructed using 3 different optimization problems. For example, problem 4 is constructed by the $Sphere$, $Rosenbrock$, $Rastrigin$, $Sphere$ and $Rosenbrock$. The two $Sphere$ are of different shifts and rotations, so are the two $Rosenbrock$. In MTO problem 8 and 9, the 5 component tasks are constructed using 5 different optimization problems.
 
\begin{table}[htb!]
	\centering
	\caption{Description of the 9 MTO test problems in test suite 2.}
\begin{tabular}{C{0.05\textwidth} L{0.06\textwidth} C{0.06\textwidth}  C{0.06\textwidth} C{0.06\textwidth} C{0.06\textwidth}}
	\toprule
	\multirow{2}{*}{ \tabincell{c}{MTO \\Problem}} & \multirow{2}{*}{ $T_1$}    & \multirow{2}{*}{ $T_2$}    & \multirow{2}{*}{$T_3$}    & \multirow{2}{*}{$T_4$}    & \multirow{2}{*}{ $T_5$} \\
	&&&&&\\
	\midrule
	
	1     & Sphere & Sphere & Sphere & Sphere & Sphere \\
	\midrule
	2     & Rosenbrock & Rosenbrock & Rosenbrock & Rosenbrock & Rosenbrock \\
	\midrule
	3     & Rastrigin & Rastrigin & Rastrigin & Rastrigin & Rastrigin \\
	\midrule
	4     & Sphere & Rosenbrock & Rastrigin & Sphere & Rosenbrock \\
	\midrule
	5     & Rastrigin & Griewank & Weierstrass & Rastrigin & Griewank \\
	\midrule
	6     & Rosenbrock & Griewank & Schwefel & Rosenbrock & Griewank \\
	\midrule
	7     & Ackley & Rastrigin & Weierstrass & Ackley & Rastrigin \\
	\midrule
	8     & Rosenbrock & Ackley & Rastrigin & Griewank & Weierstrass \\
	\midrule
	9     & Ackley & Rastrigin & Griewank & Weierstrass & Schwefel \\
	
	\bottomrule
\end{tabular}%
\label{tab:manytask_set}
\end{table}%

\subsection{Experimental Setup}

In the following experiments, the parameter settings used in the MFPSO, the SREMTO and the MFEA are according to the original literature. For the traditional PSO algorithm, the parameters are set according to Trelea \textit{et al.} \cite{trelea2003particle}. As the proposed SaMTPSO is based on the traditional PSO algorithm, hence, these settings are also employed in the SaMTPSO. The details of these settings for each of the algorithms are summarized as follow:

\begin{enumerate}
	\item Population size: $n = 50 * k$, $k$ is the number of component tasks in an MTO problem \cite{liaw2019evolutionary}
	\item Parameter settings in SaMTPSO:
	\begin{itemize}
		\item[-] based probability parameter $bp$: $0.001$
		\item[-] learning period $LP$: $10$
		\item[-] inertia weight parameter $w$: decreases linearly from 0.9 to 0.4
		\item[-] coefficient: $c_1=c_2=1.494$ in the SaMTPSO-S1, and  $c_1=c_2=c_3=1.1$ in the SaMTPSO-S2
	\end{itemize}
	\item Parameter settings in the PSO is same as the settings in the SaMTPSO.

	\item Parameter settings in MFPSO:
\begin{itemize}
	\item[-] random mating probability $rmp$: $0.3$
	\item[-] inertia weight parameter $w$: decreases linearly from 0.9 to 0.4
	\item[-] coefficient $c_1$, $c_2$ and $c_3$: $0.2$
\end{itemize}

	\item Parameter settings in SREMTO:
\begin{itemize}
	\item[-] probability for crossover: $P_\alpha = 0.7$
	\item[-] probability for mutation: $P_\beta = 1.0$
	\item[-] distribution index of SBX: $1$
	\item[-] distribution index of PM : $39$
\end{itemize}

	\item Parameter settings in MFEA:
	\begin{itemize}
		\item[-] random mating probability $rmp$: $0.3$
		\item[-] distribution index of SBX: $2$
		\item[-] distribution index of PM: $5$
	\end{itemize}

\end{enumerate}

All the following experiments are conducted for 30 runs on all algorithms. In each run, an algorithm is terminated when the maximum number of generations (denoted as maxGens) is reached. In experiment (1), (2) and (4), the maxGens is set to $2000$. In experiment (3), the maxGens is set to $10000$ for all algorithms to fully demonstrate the algorithms' performance in handling more general cases of multitasking. After the 30 runs on each algorithm, the mean and standard deviation of the achieved best objective function error values (FEVs) are presented. FEV is the difference between the objective function value of the found solution and that of the truly global optimum in each run. Besides, to fairly compare the performance of all algorithms on each MTO test problem, a popular performance score in the field of EMTO is computed based on the achieved results of each algorithm on each MTO test problem according to \cite{MFOSC2016, ding2017generalized, song2019multitasking}. More specifically, supposed there are $Q$ algorithms are comparing their achieved results on a $K$-task MTO test problem in $L$ runs, the score for an algorithm $q, q \in [1,2,...,Q]$ is computed by $score_q = \sum\limits_{j=1}^K {\sum\limits_{l=1}^L {(I_{q,j,l} - {\mu_j})/{\sigma_j}}}$, in which $I_{q,j,l}$ is algorithm $q$'s achieved FEVs on the $j$-th task in $l$-th run, ${\mu _j}$ and ${\sigma _j}$ are the mean and the standard deviation of the achieved FEVs by all algorithms in all runs. A better performance will give the algorithm a smaller score according to this definition.

 \begin{table}[htbp!]
	\centering
	\caption{Comparison of SaMTPSO (SaMTPSO-S1) and PSO in terms of the means, bracketed standard deviations and scores of the best achieved FEVs over 30 runs on test suite 1. The better results are shown in bold.}
	\resizebox{250pt}{110pt}{
		\begin{tabular}{C{0.04\textwidth} C{0.03\textwidth} C{0.12\textwidth} C{0.06\textwidth}C{0.0001\textwidth} C{0.12\textwidth}C{0.05\textwidth}}
			\toprule
			\multirow{2}[3]{*}{Problem} & \multirow{2}[3]{*}{Task} & \multicolumn{2}{c}{SaMTPSO-S1} &       & \multicolumn{2}{c}{PSO} \\
			\cmidrule{3-4}\cmidrule{6-7}          &       & Mean(Std) & Score &       & Mean(Std) & Score \\
			
			\midrule

			\multicolumn{1}{c}{\multirow{2}[2]{*}{1}} & \multicolumn{1}{l}{$T_1$} & 6.00E-3(7.70E-3) & \multicolumn{1}{c}{\multirow{2}[2]{*}{\textbf{-3.54E+1}}} &       & 1.13E-2(1.03E-2) & \multicolumn{1}{c}{\multirow{2}[2]{*}{3.54E+1}} \\
			& \multicolumn{1}{l}{$T_2$} & 2.59E+1(2.20E+1) &       &       & 3.21E+2(9.54E+1) &  \\
			\midrule
			\multicolumn{1}{c}{\multirow{2}[2]{*}{2}} & \multicolumn{1}{l}{$T_1$} & 2.79E+0(6.59E-1) & \multicolumn{1}{c}{\multirow{2}[2]{*}{\textbf{-3.86E+1}}} &       & 3.47E+0(9.11E-1) & \multicolumn{1}{c}{\multirow{2}[2]{*}{3.86E+1}} \\
			& \multicolumn{1}{l}{$T_2$} & 4.54E+1(1.72E+1) &       &       & 3.15E+2(9.27E+1) &  \\
			\midrule
			\multicolumn{1}{c}{\multirow{2}[2]{*}{3}} & \multicolumn{1}{l}{$T_1$} & 4.90E-2(6.78E-2) & \multicolumn{1}{c}{\multirow{2}[2]{*}{\textbf{-1.52E+1}}} &       & 6.43E+0(9.81E+0) & \multicolumn{1}{c}{\multirow{2}[2]{*}{1.52E+1}} \\
			& \multicolumn{1}{l}{$T_2$} & 5.45E+0(2.26E+1) &       &       & 1.63E+1(8.69E+1) &  \\
			\midrule
			\multicolumn{1}{c}{\multirow{2}[2]{*}{4}} & \multicolumn{1}{l}{$T_1$} & 3.66E+2(1.37E+2) & \multicolumn{1}{c}{\multirow{2}[2]{*}{\textbf{-1.02E+1}}} &       & 3.27E+2(9.19E+1) & \multicolumn{1}{c}{\multirow{2}[2]{*}{1.02E+1}} \\
			& \multicolumn{1}{l}{$T_2$} & 4.37E-5(1.45E-4) &       &       & 2.79E-3(3.30E-3) &  \\
			\midrule
			\multicolumn{1}{c}{\multirow{2}[2]{*}{5}} & \multicolumn{1}{l}{$T_1$} & 1.06E+0(9.51E-1) & \multicolumn{1}{c}{\multirow{2}[2]{*}{\textbf{-3.38E+1}}} &       & 3.26E+0(7.39E-1) & \multicolumn{1}{c}{\multirow{2}[2]{*}{3.38E+1}} \\
			& \multicolumn{1}{l}{$T_2$} & 9.78E+1(2.85E+1) &       &       & 2.93E+2(3.86E+2) &  \\
			\midrule
			\multicolumn{1}{c}{\multirow{2}[2]{*}{6}} & \multicolumn{1}{l}{$T_1$} & 4.14E+0(9.21E-1) & \multicolumn{1}{c}{\multirow{2}[2]{*}{\textbf{-1.32E+1}}} &       & 3.44E+0(9.36E-1) & \multicolumn{1}{c}{\multirow{2}[2]{*}{1.32E+1}} \\
			& \multicolumn{1}{l}{$T_2$} & 3.12E+0(1.59E+0) &       &       & 9.53E+0(3.08E+0) &  \\
			\midrule
			\multicolumn{1}{c}{\multirow{2}[2]{*}{7}} & \multicolumn{1}{l}{$T_1$} & 1.18E+2(4.29E+1) & \multicolumn{1}{c}{\multirow{2}[2]{*}{\textbf{-4.38E+1}}} &       & 2.21E+2(9.19E+1) & \multicolumn{1}{c}{\multirow{2}[2]{*}{4.38E+1}} \\
			& \multicolumn{1}{l}{$T_2$} & 3.38E+1(1.22E+1) &       &       & 3.01E+2(1.02E+2) &  \\
			\midrule
			\multicolumn{1}{c}{\multirow{2}[2]{*}{8}} & \multicolumn{1}{l}{$T_1$} & 7.92E-3(9.54E-3) & \multicolumn{1}{c}{\multirow{2}[2]{*}{\textbf{-2.55E+1}}} &       & 7.39E-3(6.50E-3) & \multicolumn{1}{c}{\multirow{2}[2]{*}{2.55E+1}} \\
			& \multicolumn{1}{l}{$T_2$} & 1.76E+1(2.35E+0) &       &       & 3.28E+1(5.11E+0) &  \\
			\midrule
			\multicolumn{1}{c}{\multirow{2}[2]{*}{9}} & \multicolumn{1}{l}{$T_1$} & 4.04E+2(1.18E+2) & \multicolumn{1}{c}{\multirow{2}[2]{*}{2.01E+1}} &       & 2.92E+2(9.31E+1) & \multicolumn{1}{c}{\multirow{2}[2]{*}{\textbf{-2.01E+1}}} \\
			& \multicolumn{1}{l}{$T_2$} & 3.09E+2(1.06E+3) &       &       & 3.64E-1(6.59E-1) &  \\
			\midrule
			Mean  &       & -     & \textbf{-2.17E+1} &       & -     & 2.17E+1 \\
			
			\bottomrule
		\end{tabular}%
	}
	\label{tab:benchmarkset:kt_v1}%
\end{table}%

\subsection{Experimental Results}

\subsubsection{The effectiveness, efficiency and adaptation property of knowledge transfer in SaMTPSO}

\begin{figure*}[htb!]
	\centering
	\includegraphics[width=0.9\textwidth]{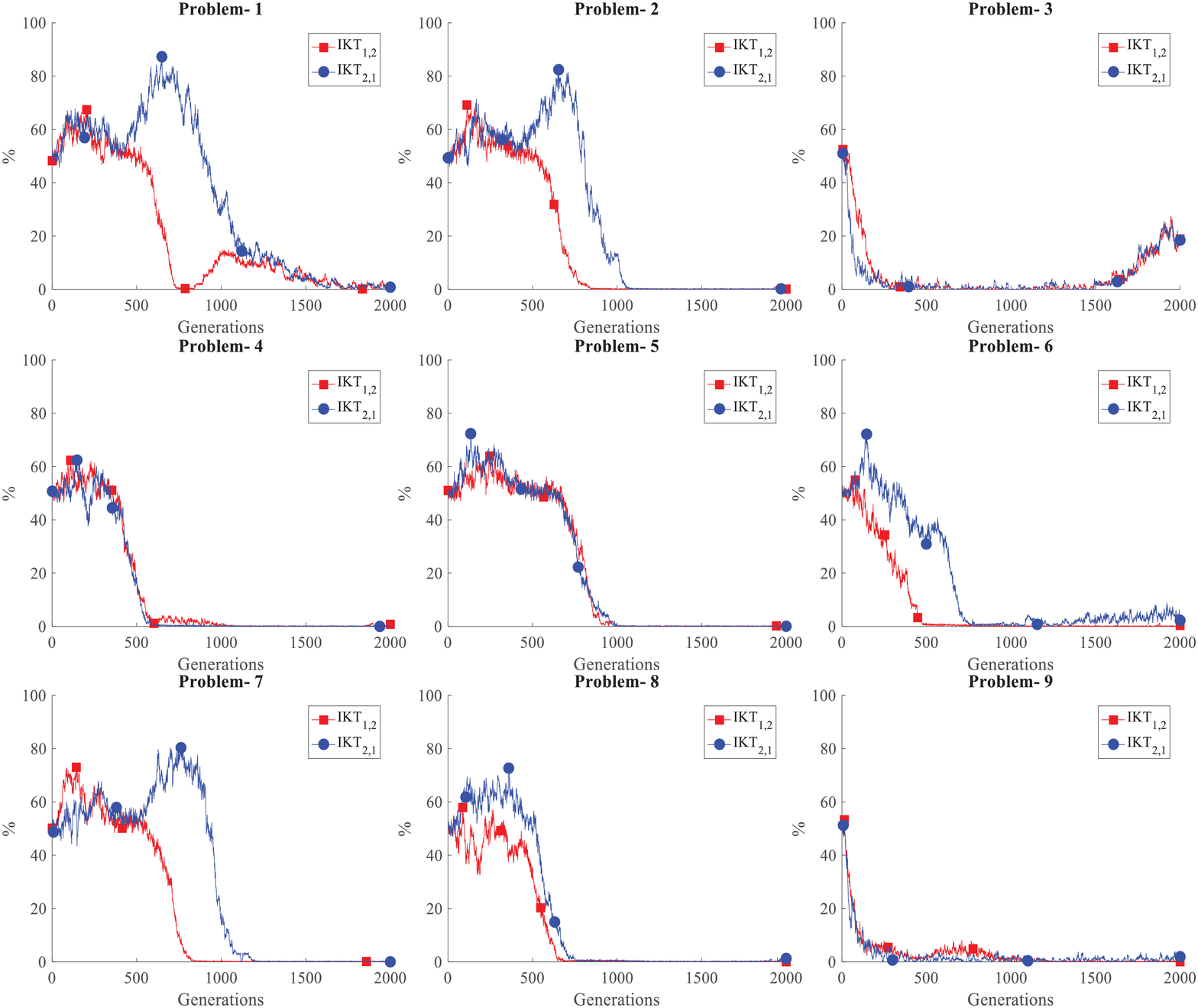}
	\caption{The adaptation details of inter-task knowledge transfer in the SaMTPSO-S1. $ITK_{1,2}$ denotes the number percentage of $T_1$'s individuals that chooses task $T_2$ as knowledge source in each generation, while $ITK_{2,1}$ denotes that of task $T_2$'s individuals that chooses task $T_1$ as knowledge source. }
	\label{Fig:KT_AvgTrend}
\end{figure*}

This experiment studies the effectiveness, efficiency and adaptation performance of the proposed SaMTPSO algorithm. The experiment is conducted by comparing the experimental results of the SaMTPSO to that of the PSO on the 9 well-designed MTO problems of test suite 1 as shown in Table \ref{tab:benchmarkset}. The two versions of the SaMTPSO algorithms are considered, each independently comparing to the PSO. In Table \ref{tab:benchmarkset:kt_v1} and Table \ref{tab:benchmarkset:kt_v2}, the means and bracketed standard deviations of the FEVs in 30 runs are reported together with the scores that are achieved on each MTO test problem by the SaMTPSO and the PSO. At the bottom of the Tables, the mean of the scores is computed to compare the overall performance of the algorithms. In Table \ref{tab:KT_AvgRate}, the averaged number percentage of a subpopulation's individuals (particles) that choose different knowledge sources over all generations and all runs, is reported to show that the knowledge transfer in the proposed SaMTPSO adapts to the inter-task relatedness. Fig. \ref{Fig:KT_AvgTrend} shows the curves of the averaged number percentage of a subpopulation's individuals that choose different knowledge sources in each generation over 30 runs in the SaMTPSO-S1. Note that, only the number percentage with respect to inter-task knowledge transfer ($ITK$) are presented, i.e., $IKT_{1,2}$ and $IKT_{2,1}$. $IKT_{1,2}$ denotes the averaged number percentage of task $T_1$'s individuals choosing task $T_2$ from $T_1$'s pool as their knowledge sources. And $IKT_{2,1}$ denotes the percentage of $T_2$'s individuals choosing $T_1$ as their knowledge sources. The results on SaMTPSO-S2 is similar to that of the SaMTPSO-S1. Therefore, only the results on the SaMTPSO-S1 are presented here due to the space limitation of paper.

\begin{table}[htbp!]
	\centering
	\caption{Comparison of SaMTPSO (SaMTPSO-S2) and PSO in terms of the means, bracketed standard deviations and scores of the best achieved FEVs over 30 runs on test suite 1. The better results are shown in bold.}
	\resizebox{250pt}{110pt}{
		\begin{tabular}{C{0.04\textwidth} C{0.03\textwidth} C{0.12\textwidth} C{0.06\textwidth}C{0.0001\textwidth} C{0.12\textwidth}C{0.05\textwidth}}
			\toprule
			\multirow{2}[3]{*}{Problem} & \multirow{2}[3]{*}{Task} & \multicolumn{2}{c}{SaMTPSO-S2} &       & \multicolumn{2}{c}{PSO} \\
			\cmidrule{3-4}\cmidrule{6-7}          &       & Mean(Std) & Score &       & Mean(Std) & Score \\
			
			\midrule
			
			\multicolumn{1}{c}{\multirow{2}[2]{*}{1}} & \multicolumn{1}{l}{$T_1$} & 6.19E-3(7.55E-3) & \multirow{2}[2]{*}{\textbf{-3.54E+1}} &       & 1.13E-2(1.03E-2) & \multirow{2}[2]{*}{3.54E+1} \\
			& \multicolumn{1}{l}{$T_2$} & 1.41E+1(1.80E+1) &       &       & 3.21E+2(9.54E+1) &  \\
			\midrule
			\multicolumn{1}{c}{\multirow{2}[2]{*}{2}} & \multicolumn{1}{l}{$T_1$} & 2.87E+0(5.91E-1) & \multirow{2}[2]{*}{\textbf{-3.78E+1}} &       & 3.47E+0(9.11E-1) & \multirow{2}[2]{*}{3.78E+1} \\
			& \multicolumn{1}{l}{$T_2$} & 4.29E+1(1.43E+1) &       &       & 3.15E+2(9.27E+1) &  \\
			\midrule
			\multicolumn{1}{c}{\multirow{2}[2]{*}{3}} & \multicolumn{1}{l}{$T_1$} & 6.91E-2(1.31E-1) & \multirow{2}[2]{*}{\textbf{-1.65E+1}} &       & 6.43E+0(9.81E+0) & \multirow{2}[2]{*}{1.65E+1} \\
			& \multicolumn{1}{l}{$T_2$} & 2.88E-1(8.23E-1) &       &       & 1.63E+1(8.69E+1) &  \\
			\midrule
			\multicolumn{1}{c}{\multirow{2}[2]{*}{4}} & \multicolumn{1}{l}{$T_1$} & 3.07E+2(9.09E+1) & \multirow{2}[2]{*}{\textbf{-1.87E+1}} &       & 3.27E+2(9.19E+1) & \multirow{2}[2]{*}{1.87E+1} \\
			& \multicolumn{1}{l}{$T_2$} & 6.86E-7(2.40E-6) &       &       & 2.79E-3(3.30E-3) &  \\
			\midrule
			\multicolumn{1}{c}{\multirow{2}[2]{*}{5}} & \multicolumn{1}{l}{$T_1$} & 1.17E+0(8.41E-1) & \multirow{2}[2]{*}{\textbf{-3.38E+1}} &       & 3.26E+0(7.39E-1) & \multirow{2}[2]{*}{3.38E+1} \\
			& \multicolumn{1}{l}{$T_2$} & 1.01E+2(2.77E+1) &       &       & 2.93E+2(3.86E+2) &  \\
			\midrule
			\multicolumn{1}{c}{\multirow{2}[2]{*}{6}} & \multicolumn{1}{l}{$T_1$} & 3.25E+0(6.93E-1) & \multirow{2}[2]{*}{\textbf{-2.91E+1}} &       & 3.44E+0(9.36E-1) & \multirow{2}[2]{*}{2.91E+1} \\
			& \multicolumn{1}{l}{$T_2$} & 2.11E+0(7.01E-1) &       &       & 9.53E+0(3.08E+0) &  \\
			\midrule
			\multicolumn{1}{c}{\multirow{2}[2]{*}{7}} & \multicolumn{1}{l}{$T_1$} & 1.28E+2(6.23E+1) & \multirow{2}[2]{*}{\textbf{-4.15E+1}} &       & 2.21E+2(9.19E+1) & \multirow{2}[2]{*}{4.15E+1} \\
			& \multicolumn{1}{l}{$T_2$} & 3.20E+1(2.32E+1) &       &       & 3.01E+2(1.02E+2) &  \\
			\midrule
			\multicolumn{1}{c}{\multirow{2}[2]{*}{8}} & \multicolumn{1}{l}{$T_1$} & 8.43E-3(8.52E-3) & \multirow{2}[2]{*}{\textbf{-2.36E+1}} &       & 7.39E-3(6.50E-3) & \multirow{2}[2]{*}{2.36E+1} \\
			& \multicolumn{1}{l}{$T_2$} & 1.94E+1(2.28E+0) &       &       & 3.28E+1(5.11E+0) &  \\
			\midrule
			\multicolumn{1}{c}{\multirow{2}[2]{*}{9}} & \multicolumn{1}{l}{$T_1$} & 3.59E+2(9.68E+1) & \multirow{2}[2]{*}{1.40E+1} &       & 2.92E+2(9.31E+1) & \multirow{2}[2]{*}{\textbf{-1.40E+1}} \\
			& \multicolumn{1}{l}{$T_2$} & 1.39E+1(7.11E+1) &       &       & 3.64E-1(6.59E-1) &  \\
			\midrule
			Mean  &       & -     & \textbf{-2.47E+1} &       & -     & 2.47E+1 \\
			
			\bottomrule
		\end{tabular}%
	}
	\label{tab:benchmarkset:kt_v2}%
\end{table}%

In Table \ref{tab:benchmarkset:kt_v1}, the proposed SaMTPSO-S1 achieves better scores than the corresponding PSO algorithm, with a mean score $-2.17E+1$ comparing to the $2.17E+1$ of the PSO. On 8 of the 9 MTO problems, the SaMTPSO-S1 performs better than the PSO algorithm. Only on the problem 9, the SaMTPSO-S1 achieves slightly inefficient results due to the very small inter-task relatedness as given in Table \ref{tab:benchmarkset}. Similarly, the proposed SaMTPSO-S2 achieves better results comparing to the PSO algorithm as shown in Table \ref{tab:benchmarkset:kt_v2}. Therefore, the proposed SaMTPSO is efficient in handling MTO problems.

Table \ref{tab:KT_AvgRate} presents the averaged number percentage of a subpopulation's individuals that chooses different knowledge sources in all generations across all runs. For example, in MTO problem 1, there are average $79.58\%$ individuals in the subpopulation of task $T_1$ who choose task $T_1$ from the pool as their knowledge sources at one generation in the SaMTPSO-S1. And the rest $20.42\%$ individuals choose $T_2$ as their knowledge source, which is called inter-task knowledge transfer. From this table, we can find that, on MTO problems with very small inter-task relatedness, such as problem 3, 6 and 9 according to the $Rs$ in Table \ref{tab:benchmarkset}, the number percentage of individuals regarding to inter-task knowledge transfer in both the SaMTPSO-S1 and SaMTPSO-S2 are usually in a very low level, about $3\%-10\%$. On the rest MTO problems, these percentages are usually in the range of $10\%-35\%$ due to the higher inter-task relatedness between the component tasks of the MTO problems. Therefore, knowledge transfer in the proposed SaMTPSO adapts to task relatedness in overall.

\begin{table}[htbp]
	\centering
	\caption{The averaged number percentage of a subpopulation's individuals that choose different knowledge sources in all generations across all runs. The ones with respect to inter-task knowledge transfer are shown in bold.}
	\begin{tabular}{C{0.04\textwidth} C{0.04\textwidth} C{0.06\textwidth} C{0.06\textwidth} C{0.00001\textwidth}C{0.06\textwidth} C{0.06\textwidth}}
		\toprule
		
		\multirow{3}[6]{*}{Problem} & \multirow{3}[6]{*}{Tasks} & \multicolumn{2}{c}{SaMTPSO-S1} &       & \multicolumn{2}{c}{SaMTPSO-S2} \\
		\cmidrule{3-4}\cmidrule{6-7}          &       & \multicolumn{2}{c}{Knowledge Source} &       & \multicolumn{2}{c}{Knowledge Source} \\
		\cmidrule{3-4}\cmidrule{6-7}          &       & $T_1$    & $T_2$    &       & $T_1$    & $T_2$ \\
		\midrule
		
		\multirow{2}[2]{*}{1} & $T_1$    & 79.58\% & \textbf{20.42\%} &       & 77.76\% & \textbf{22.24\%} \\
		& $T_2$    & \textbf{34.44\%} & 65.56\% &       & \textbf{33.08\%} & 66.92\% \\
		\midrule
		\multirow{2}[2]{*}{2} & $T_1$    & 82.35\% & \textbf{17.65\%} &       & 80.28\% & \textbf{19.72\%} \\
		& $T_2$    & \textbf{27.13\%} & 72.87\% &       & \textbf{29.22\%} & 70.78\% \\
		\midrule
		\multirow{2}[2]{*}{3} & $T_1$    & 94.19\% & \textbf{5.81\%} &       & 91.30\% & \textbf{8.70\%} \\
		& $T_2$    & \textbf{4.81\%} & 95.19\% &       & \textbf{6.52\%} & 93.48\% \\
		\midrule
		\multirow{2}[2]{*}{4} & $T_1$    & 87.12\% & \textbf{12.88\%} &       & 87.98\% & \textbf{12.02\%} \\
		& $T_2$    & \textbf{11.88\%} & 88.12\% &       & \textbf{7.03\%} & 92.97\% \\
		\midrule
		\multirow{2}[2]{*}{5} & $T_1$    & 79.10\% & \textbf{20.90\%} &       & 80.42\% & \textbf{19.58\%} \\
		& $T_2$    & \textbf{21.59\%} & 78.41\% &       & \textbf{21.89\%} & 78.11\% \\
		\midrule
		\multirow{2}[2]{*}{6} & $T_1$    & 92.54\% & \textbf{7.46\%} &       & 93.47\% & \textbf{6.53\%} \\
		& $T_2$    & \textbf{16.49\%} & 83.51\% &       & \textbf{16.65\%} & 83.35\% \\
		\midrule
		\multirow{2}[2]{*}{7} & $T_1$    & 80.40\% & \textbf{19.60\%} &       & 80.01\% & \textbf{19.99\%} \\
		& $T_2$    & \textbf{30.02\%} & 69.98\% &       & \textbf{27.53\%} & 72.47\% \\
		\midrule
		\multirow{2}[2]{*}{8} & $T_1$    & 87.51\% & \textbf{12.49\%} &       & 92.89\% & \textbf{7.11\%} \\
		& $T_2$    & \textbf{17.43\%} & 82.57\% &       & \textbf{17.94\%} & 82.06\% \\
		\midrule
		\multirow{2}[2]{*}{9} & $T_1$    & 96.82\% & \textbf{3.18\%} &       & 93.65\% & \textbf{6.35\%} \\
		& $T_2$    & \textbf{2.55\%} & 97.45\% &       & \textbf{3.02\%} & 96.98\% \\

		\bottomrule
	\end{tabular}%
	\label{tab:KT_AvgRate}%
\end{table}%

From Fig. \ref{Fig:KT_AvgTrend}, we can observe the adaptation details of inter-task knowledge transfer in the proposed SaMTPSO via the $ITK_{1,2}$ and the $ITK_{2,1}$. For all the 9 MTO problems, there are about $50\%$ individuals in each subpopulation involving inter-task knowledge transfer in the early stage. Two reasons should be considered, 1) the knowledge sources in each pool are initialized with the same chosen probabilities; 2) it is usually easy to find better solutions for the component tasks of an MTO problem at the beginning of optimization, which makes it no difference in choosing which kind of knowledge sources. As the optimization goes, these number percentages keeps for a while on the MTO problems with high inter-task relatedness according to the $Rs$ in Table \ref{tab:benchmarkset}, such as problem 1, 2, 4, 5, 7 and 8. On problem 3 and 9, these number percentages drop very soon at the early stage of optimization due to the lack of inter-task relatedness ($Rs = 0$ as given in Table \ref{tab:benchmarkset}). Problem 6 is an exception, which may be because the component tasks of problem 6 are of different dimensions in comparing to that of the problem 3 and 9. At the last stage of optimization, there are nearly no inter-task knowledge transfer in most of the MTO problems, because the component tasks in these MTO problems are of different optimal solutions, which leads to the failure of inter-task knowledge transfer. For problem 1 and 2, the SaMTPSO-S1 may probably have only found local optimal solutions for the component tasks, though they have the same global optimal solutions. Therefore, these percentages regarding to inter-task knowledge transfer are still nearly equal to 0. For problem 3, the SaMTPSO-S1 have found good-enough solutions for the two component tasks as shown in Table \ref{tab:benchmarkset:kt_v1}, achieving an averaged FEV $4.90E-2$ on task $T_1$ and $5.45E+0$ on $T_2$. As the component tasks have the same global optimal solutions, inter-task knowledge transfer can success more easily. As a result, the percentages regarding to inter-task knowledge transfer are increasing at the end of optimization as shown in the figure.

\subsubsection{Comparison of SaMTPSO with MFPSO, MFEA and SREMTO on test suite 1}

In this experiment, the experimental results of the SaMTPSO-S1 and the SaMTPSO-S2 on test suite 1 are compared to that of the MFPSO, the SREMTO and the MFEA. In Table \ref{tab:cmp_simple_2tasks}, the means, bracketed standard deviations of the FEVs over 30 runs are reported along with the achieved scores by all these algorithms.
 \begin{table*}[!htbp]
	\centering
	\caption{Comparison of SaMTPSO-S1, SaMTPSO-S2, MFPSO, SREMTO and MFEA in terms of the means, bracketed standard deviations of the best achieved FEVs over 30 runs and the achieved scores on the 9 MTO problems in test suite 1. The better results are shown in bold.}
	\resizebox{516pt}{120pt}{
		\begin{tabular}{C{0.04\textwidth} C{0.04\textwidth} C{0.13\textwidth}C{0.06\textwidth}C{0.0000001\textwidth}  C{0.13\textwidth}C{0.06\textwidth}C{0.0000001\textwidth} C{0.13\textwidth}C{0.06\textwidth}C{0.0000001\textwidth} C{0.13\textwidth}C{0.06\textwidth}C{0.0000001\textwidth} C{0.13\textwidth}C{0.06\textwidth}C{0.0000001\textwidth}}
			\toprule
			
			\multirow{2}[4]{*}{\tabincell{c}{Problem \\ Sets\\}} & \multirow{2}[4]{*}{Task} & \multicolumn{2}{c}{SaMTPSO-S1} &       & \multicolumn{2}{c}{SaMTPSO-S2} &       & \multicolumn{2}{c}{MFPSO} &       & \multicolumn{2}{c}{SREMTO} &       & \multicolumn{2}{c}{MFEA} \\
			\cmidrule{3-4}\cmidrule{6-7}\cmidrule{9-10}\cmidrule{12-13}\cmidrule{15-16}          &       & Mean(Std) & Score &       & Mean(Std) & Score &       & Mean(Std) & Score &       & Mean(Std) & Score &       & Mean(Std) & Score \\
			\midrule
			
			\multirow{2}[2]{*}{1} & $T_1$    & 6.00E-3(7.70E-3) & \multirow{2}[2]{*}{-3.61E+1} &       & 6.19E-3(7.55E-3) & \multirow{2}[2]{*}{\textbf{-3.85E+1}} &       & 9.13E-1(1.12E-1) & \multirow{2}[2]{*}{1.12E+2} &       & 4.67E-3(1.00E-2) & \multirow{2}[2]{*}{-3.86E+1} &       & 8.68E-2(2.51E-2) & \multirow{2}[2]{*}{1.18E+0} \\
			& $T_2$    & 2.59E+1(2.20E+1) &       &       & 1.41E+1(1.80E+1) &       &       & 3.74E+2(2.27E+1) &       &       & 1.44E+1(2.14E+1) &       &       & 1.72E+2(4.49E+1) &  \\
			\midrule
			\multirow{2}[2]{*}{2} & $T_1$    & 2.79E+0(6.59E-1) & \multirow{2}[2]{*}{\textbf{-4.20E+1}} &       & 2.87E+0(5.91E-1) & \multirow{2}[2]{*}{-4.11E+1} &       & 7.08E+0(8.90E-1) & \multirow{2}[2]{*}{1.06E+2} &       & 3.38E+0(1.09E+0) & \multirow{2}[2]{*}{-2.83E+1} &       & 4.28E+0(6.52E-1) & \multirow{2}[2]{*}{5.13E+0} \\
			& $T_2$    & 4.54E+1(1.72E+1) &       &       & 4.29E+1(1.43E+1) &       &       & 5.06E+2(5.68E+1) &       &       & 6.83E+1(3.91E+1) &       &       & 1.80E+2(6.15E+1) &  \\
			\midrule
			\multirow{2}[2]{*}{3} & $T_1$    & 4.90E-2(6.78E-2) & \multirow{2}[2]{*}{\textbf{-6.20E+1}} &       & 6.91E-2(1.31E-1) & \multirow{2}[2]{*}{\textbf{-6.20E+1}} &       & 2.12E+1(3.41E-2) & \multirow{2}[2]{*}{8.10E+1} &       & 2.09E+1(3.84E-1) & \multirow{2}[2]{*}{2.90E+1} &       & 2.01E+1(5.47E-2) & \multirow{2}[2]{*}{1.40E+1} \\
			& $T_2$    & 5.45E+0(2.26E+1) &       &       & 2.88E-1(8.23E-1) &       &       & 1.39E+4(1.07E+3) &       &       & 5.09E+3(6.04E+2) &       &       & 2.93E+3(4.54E+2) &  \\
			\midrule
			\multirow{2}[2]{*}{4} & $T_1$    & 3.66E+2(1.37E+2) & \multirow{2}[2]{*}{-2.76E+1} &       & 3.07E+2(9.09E+1) & \multirow{2}[2]{*}{-3.46E+1} &       & 8.95E+2(9.26E+1) & \multirow{2}[2]{*}{1.08E+2} &       & 2.59E+2(7.80E+1) & \multirow{2}[2]{*}{\textbf{-4.03E+1}} &       & 5.49E+2(9.38E+1) & \multirow{2}[2]{*}{-5.85E+0} \\
			& $T_2$    & 4.37E-5(1.45E-4) &       &       & 6.86E-7(2.40E-6) &       &       & 4.62E+3(8.64E+2) &       &       & 1.28E-15(6.31E-15) &       &       & 3.63E-1(1.12E-1) &  \\
			\midrule
			\multirow{2}[2]{*}{5} & $T_1$    & 1.06E+0(9.51E-1) & \multirow{2}[2]{*}{\textbf{-3.69E+1}} &       & 1.17E+0(8.41E-1) & \multirow{2}[2]{*}{-3.54E+1} &       & 6.75E+0(1.00E+0) & \multirow{2}[2]{*}{1.01E+2} &       & 2.28E+0(6.06E-1) & \multirow{2}[2]{*}{-2.06E+1} &       & 3.17E+0(5.60E-1) & \multirow{2}[2]{*}{-8.48E+0} \\
			& $T_2$    & 9.78E+1(2.85E+1) &       &       & 1.01E+2(2.77E+1) &       &       & 1.21E+5(7.49E+4) &       &       & 8.71E+1(3.46E+1) &       &       & 2.20E+2(4.56E+1) &  \\
			\midrule
			\multirow{2}[2]{*}{6} & $T_1$    & 4.14E+0(9.21E-1) & \multirow{2}[2]{*}{-4.01E+1} &       & 3.25E+0(6.93E-1) & \multirow{2}[2]{*}{\textbf{-4.85E+1}} &       & 1.17E+1(1.32E+0) & \multirow{2}[2]{*}{2.46E+1} &       & 3.75E+0(6.10E-1) & \multirow{2}[2]{*}{-4.00E+1} &       & 1.99E+1(7.64E-2) & \multirow{2}[2]{*}{1.04E+2} \\
			& $T_2$    & 3.12E+0(1.59E+0) &       &       & 2.11E+0(7.01E-1) &       &       & 1.02E+1(1.56E+0) &       &       & 3.55E+0(1.25E+0) &       &       & 1.99E+1(3.79E+0) &  \\
			\midrule
			\multirow{2}[2]{*}{7} & $T_1$    & 1.18E+2(4.29E+1) & \multirow{2}[2]{*}{-3.46E+1} &       & 1.28E+2(6.23E+1) & \multirow{2}[2]{*}{\textbf{-3.48E+1}} &       & 3.69E+5(1.51E+5) & \multirow{2}[2]{*}{1.07E+2} &       & 8.75E+1(3.76E+1) & \multirow{2}[2]{*}{-3.07E+1} &       & 2.57E+2(5.67E+1) & \multirow{2}[2]{*}{-7.24E+0} \\
			& $T_2$    & 3.38E+1(1.22E+1) &       &       & 3.20E+1(2.32E+1) &       &       & 5.66E+2(1.16E+2) &       &       & 6.22E+1(2.91E+1) &       &       & 2.31E+2(1.01E+2) &  \\
			\midrule
			\multirow{2}[2]{*}{8} & $T_1$    & 7.92E-3(9.54E-3) & \multirow{2}[2]{*}{\textbf{-4.48E+1}} &       & 8.43E-3(8.52E-3) & \multirow{2}[2]{*}{-3.35E+1} &       & 1.09E+0(3.29E-2) & \multirow{2}[2]{*}{9.96E+1} &       & 8.53E-3(9.04E-3) & \multirow{2}[2]{*}{-3.19E+1} &       & 9.57E-2(1.91E-2) & \multirow{2}[2]{*}{1.06E+1} \\
			& $T_2$    & 1.76E+1(2.35E+0) &       &       & 1.94E+1(2.28E+0) &       &       & 2.85E+1(1.01E+0) &       &       & 1.97E+1(3.08E+0) &       &       & 2.55E+1(2.71E+0) &  \\
			\midrule
			\multirow{2}[2]{*}{9} & $T_1$    & 4.04E+2(1.18E+2) & \multirow{2}[2]{*}{-3.82E+1} &       & 3.59E+2(9.68E+1) & \multirow{2}[2]{*}{\textbf{-4.28E+1}} &       & 1.50E+3(2.16E+2) & \multirow{2}[2]{*}{1.11E+2} &       & 2.70E+2(4.76E+1) & \multirow{2}[2]{*}{-1.93E+1} &       & 5.96E+2(1.27E+2) & \multirow{2}[2]{*}{-1.06E+1} \\
			& $T_2$    & 3.09E+2(1.06E+3) &       &       & 1.39E+1(7.11E+1) &       &       & 1.37E+4(1.42E+3) &       &       & 4.97E+3(5.45E+2) &       &       & 2.92E+3(3.73E+2) &  \\
			\midrule
			\multicolumn{2}{c}{Mean} & -     & -4.03E+1 &       & -     & \textbf{-4.12E+1} &       & -     & 9.46E+1 &       & -     & -2.45E+1 &       & -     & 1.14E+1 \\

			\bottomrule
		\end{tabular}%
	}
	\label{tab:cmp_simple_2tasks}%
\end{table*}%

Table \ref{tab:cmp_simple_2tasks} shows the superiority of the proposed SaMTPSO via the SaMTPSO-S1 and SaMTPSO-S2. Comparing to the MFPSO, the SREMTO and the MFEA, the SaMTPSO-S1 and the SaMTPSO-S2 have achieved the best mean scores $-4.03E+1$ and $-4.12E+1$ respectively. On 4 of the 9 MTO problems, the SaMTPSO-S1 achieves the best scores. On 5 of the 9 MTO problems, the SaMTPSO-S2 achieves the best scores. Therefore, the proposed SaMTPSO is promising in handling MTO problems.

\subsubsection{Comparison of SaMTPSO-S1, SaMTPSO-S2, PSO, SREMTO and MFEA on test suite 2}

This experiment studies the performance of the proposed SaMTPSO on handling MTO problems with more than two component tasks via the comparison of experimental results of the SaMTPSO-S1, the SaMTPSO-S2, the PSO, the SREMTO and the MFEA on the 9 MTO 5-task MTO test problems as shown in Table \ref{tab:manytask_set}. To handle such kind of MTO problems, the MFPSO need further study and improvement. Therefore, the MFPSO is not included in this experiment. In Table \ref{tab:cmp_many_5Tasks}, the means, bracketed standard deviations of the FEVs over 30 runs are reported along with the achieved scores by all these algorithms. 
 
\begin{table*}[htbp!]
	\centering
\caption{Comparison of SaMTPSO-S1, SaMTPSO-S2, MFPSO, SREMTO and MFEA in terms of the means, bracketed standard deviations of the best achieved FEVs over 30 runs and the achieved scores on 9 MTO problems in test suite 2. The better results are shown in bold.}
\resizebox{516pt}{190pt}{
	\begin{tabular}{C{0.04\textwidth} C{0.04\textwidth} C{0.1\textwidth}C{0.06\textwidth}C{0.00001\textwidth}C{0.1\textwidth}C{0.06\textwidth}C{0.00001\textwidth}C{0.1\textwidth}C{0.06\textwidth}C{0.00001\textwidth}C{0.1\textwidth}C{0.06\textwidth}C{0.00001\textwidth}C{0.1\textwidth}C{0.06\textwidth}}
	\toprule
    \multirow{2}[4]{*}{Problem} & \multirow{2}[4]{*}{Task} & \multicolumn{2}{c}{SaMTPSO-S1} &       & \multicolumn{2}{c}{SaMTPSO-S2} &       & \multicolumn{2}{c}{PSO} &       & \multicolumn{2}{c}{SREMTO} &       & \multicolumn{2}{c}{MFEA} \\
\cmidrule{3-4}\cmidrule{6-7}\cmidrule{9-10}\cmidrule{12-13}\cmidrule{15-16}          &       & Mean(Std) & \multicolumn{1}{c}{Score} &       & Mean(Std) & \multicolumn{1}{c}{Score} &       & Mean(Std) & \multicolumn{1}{c}{Score} &       & Mean(Std) & \multicolumn{1}{c}{Score} &       & Mean(Std) & \multicolumn{1}{c}{Score} \\

\cmidrule{1-4}\cmidrule{6-7}\cmidrule{9-10}\cmidrule{12-13}\cmidrule{15-16}    \multirow{5}[2]{*}{1} & $T_1$    & \multicolumn{1}{l}{2.62E-8(9.42E-8)} & \multicolumn{1}{c}{\multirow{5}[2]{*}{\textbf{ -6.79E+1 }}} &       & \multicolumn{1}{l}{5.14E-7(2.76E-6)} & \multicolumn{1}{c}{\multirow{5}[2]{*}{ -6.78E+1 }} &       & \multicolumn{1}{l}{5.66E-22(2.76E-21)} & \multicolumn{1}{c}{\multirow{5}[2]{*}{\textbf{-6.79E+1}}} &       & \multicolumn{1}{l}{5.06E-27(7.18E-28)} & \multicolumn{1}{c}{\multirow{5}[2]{*}{\textbf{-6.79E+1}}} &       & \multicolumn{1}{l}{2.67E-3(1.23E-3)} & \multicolumn{1}{c}{\multirow{5}[2]{*}{2.71E+2}} \\
& $T_2$    & \multicolumn{1}{l}{8.09E-9(4.41E-8)} &       &       & \multicolumn{1}{l}{1.44E-9(5.42E-9)} &       &       & \multicolumn{1}{l}{2.19E-21(9.59E-21)} &       &       & \multicolumn{1}{l}{5.95E-27(7.45E-28)} &       &       & \multicolumn{1}{l}{2.21E-3(8.59E-4)} &  \\
& $T_3$    & \multicolumn{1}{l}{6.47E-11(2.50E-10)} &       &       & \multicolumn{1}{l}{6.40E-11(1.98E-10)} &       &       & \multicolumn{1}{l}{8.08E-21(4.22E-20)} &       &       & \multicolumn{1}{l}{5.35E-27(7.42E-28)} &       &       & \multicolumn{1}{l}{2.50E-3(1.22E-3)} &  \\
& $T_4$    & \multicolumn{1}{l}{8.59E-13(4.46E-12)} &       &       & \multicolumn{1}{l}{5.95E-7(3.24E-6)} &       &       & \multicolumn{1}{l}{2.51E-20(1.37E-19)} &       &       & \multicolumn{1}{l}{5.64E-27(7.20E-28)} &       &       & \multicolumn{1}{l}{2.23E-3(8.79E-4)} &  \\
& $T_5$    & \multicolumn{1}{l}{5.83E-12(2.01E-11)} &       &       & \multicolumn{1}{l}{8.32E-10(3.11E-9)} &       &       & \multicolumn{1}{l}{3.58E-21(1.37E-20)} &       &       & \multicolumn{1}{l}{4.61E-27(6.19E-28)} &       &       & \multicolumn{1}{l}{2.30E-3(8.39E-4)} &  \\
\midrule
\multirow{5}[2]{*}{2} & $T_1$    & \multicolumn{1}{l}{1.01E+2(8.88E+1)} & \multicolumn{1}{c}{\multirow{5}[2]{*}{ -3.23E+1 }} &       & \multicolumn{1}{l}{8.97E+1(6.40E+1)} & \multicolumn{1}{c}{\multirow{5}[2]{*}{ -3.02E+1 }} &       & \multicolumn{1}{l}{2.00E+2(4.55E+2)} & \multicolumn{1}{c}{\multirow{5}[2]{*}{-2.44E+1}} &       & \multicolumn{1}{l}{8.02E+1(7.89E+1)} & \multicolumn{1}{c}{\multirow{5}[2]{*}{\textbf{-3.26E+1}}} &       & \multicolumn{1}{l}{1.02E+3(2.06E+3)} & \multicolumn{1}{c}{\multirow{5}[2]{*}{1.19E+2}} \\
& $T_2$    & \multicolumn{1}{l}{1.44E+2(1.61E+2)} &       &       & \multicolumn{1}{l}{1.60E+2(1.81E+2)} &       &       & \multicolumn{1}{l}{1.76E+2(4.28E+2)} &       &       & \multicolumn{1}{l}{1.26E+2(1.64E+2)} &       &       & \multicolumn{1}{l}{1.51E+3(2.72E+3)} &  \\
& $T_3$    & \multicolumn{1}{l}{9.64E+1(9.10E+1)} &       &       & \multicolumn{1}{l}{1.29E+2(1.62E+2)} &       &       & \multicolumn{1}{l}{1.92E+2(4.23E+2)} &       &       & \multicolumn{1}{l}{8.23E+1(9.63E+1)} &       &       & \multicolumn{1}{l}{1.46E+3(2.48E+3)} &  \\
& $T_4$    & \multicolumn{1}{l}{1.03E+2(1.23E+2)} &       &       & \multicolumn{1}{l}{1.33E+2(3.49E+2)} &       &       & \multicolumn{1}{l}{1.60E+2(2.73E+2)} &       &       & \multicolumn{1}{l}{1.03E+2(1.41E+2)} &       &       & \multicolumn{1}{l}{1.21E+3(2.04E+3)} &  \\
& $T_5$    & \multicolumn{1}{l}{1.36E+2(1.21E+2)} &       &       & \multicolumn{1}{l}{1.53E+2(1.40E+2)} &       &       & \multicolumn{1}{l}{1.48E+2(2.36E+2)} &       &       & \multicolumn{1}{l}{1.80E+2(2.00E+2)} &       &       & \multicolumn{1}{l}{1.25E+3(2.47E+3)} &  \\
\midrule
\multirow{5}[2]{*}{3} & $T_1$    & \multicolumn{1}{l}{1.82E+2(4.67E+1)} & \multicolumn{1}{c}{\multirow{5}[2]{*}{\textbf{ -1.13E+2 }}} &       & \multicolumn{1}{l}{2.06E+2(8.01E+1)} & \multicolumn{1}{c}{\multirow{5}[2]{*}{ -8.85E+1 }} &       & \multicolumn{1}{l}{3.28E+2(1.08E+2)} & \multicolumn{1}{c}{\multirow{5}[2]{*}{3.62E+1}} &       & \multicolumn{1}{l}{1.99E+2(4.94E+1)} & \multicolumn{1}{c}{\multirow{5}[2]{*}{-7.90E+1}} &       & \multicolumn{1}{l}{6.10E+2(1.08E+2)} & \multicolumn{1}{c}{\multirow{5}[2]{*}{2.44E+2}} \\
& $T_2$    & \multicolumn{1}{l}{1.58E+2(2.99E+1)} &       &       & \multicolumn{1}{l}{1.69E+2(7.13E+1)} &       &       & \multicolumn{1}{l}{3.45E+2(9.45E+1)} &       &       & \multicolumn{1}{l}{1.96E+2(3.97E+1)} &       &       & \multicolumn{1}{l}{5.61E+2(1.33E+2)} &  \\
& $T_3$    & \multicolumn{1}{l}{1.58E+2(4.26E+1)} &       &       & \multicolumn{1}{l}{1.80E+2(4.34E+1)} &       &       & \multicolumn{1}{l}{3.35E+2(8.83E+1)} &       &       & \multicolumn{1}{l}{2.04E+2(5.50E+1)} &       &       & \multicolumn{1}{l}{5.82E+2(1.29E+2)} &  \\
& $T_4$    & \multicolumn{1}{l}{1.51E+2(3.50E+1)} &       &       & \multicolumn{1}{l}{2.06E+2(6.99E+1)} &       &       & \multicolumn{1}{l}{3.26E+2(8.98E+1)} &       &       & \multicolumn{1}{l}{1.99E+2(5.19E+1)} &       &       & \multicolumn{1}{l}{5.22E+2(1.19E+2)} &  \\
& $T_5$    & \multicolumn{1}{l}{1.71E+2(3.43E+1)} &       &       & \multicolumn{1}{l}{1.92E+2(5.87E+1)} &       &       & \multicolumn{1}{l}{3.19E+2(7.89E+1)} &       &       & \multicolumn{1}{l}{2.07E+2(3.87E+1)} &       &       & \multicolumn{1}{l}{5.48E+2(1.13E+2)} &  \\
\midrule
\multirow{5}[2]{*}{4} & $T_1$    & \multicolumn{1}{l}{1.28E-8(4.76E-8)} & \multicolumn{1}{c}{\multirow{5}[2]{*}{ -4.08E+1 }} &       & \multicolumn{1}{l}{1.83E-9(6.30E-9)} & \multicolumn{1}{c}{\multirow{5}[2]{*}{ -4.71E+1 }} &       & \multicolumn{1}{l}{1.80E-21(5.61E-21)} & \multicolumn{1}{c}{\multirow{5}[2]{*}{\textbf{-5.29E+1}}} &       & \multicolumn{1}{l}{4.50E-27(7.17E-28)} & \multicolumn{1}{c}{\multirow{5}[2]{*}{-2.07E+1}} &       & \multicolumn{1}{l}{2.53E-3(1.26E-3)} & \multicolumn{1}{c}{\multirow{5}[2]{*}{1.62E+2}} \\
& $T_2$    & \multicolumn{1}{l}{1.04E+2(2.01E+2)} &       &       & \multicolumn{1}{l}{1.22E+2(2.97E+2)} &       &       & \multicolumn{1}{l}{1.48E+2(2.02E+2)} &       &       & \multicolumn{1}{l}{6.63E+1(5.78E+1)} &       &       & \multicolumn{1}{l}{5.62E+2(1.39E+3)} &  \\
& $T_3$    & \multicolumn{1}{l}{3.86E+0(9.96E-1)} &       &       & \multicolumn{1}{l}{3.53E+0(6.81E-1)} &       &       & \multicolumn{1}{l}{3.46E+0(9.45E-1)} &       &       & \multicolumn{1}{l}{4.71E+0(1.42E+0)} &       &       & \multicolumn{1}{l}{4.84E+0(9.83E-1)} &  \\
& $T_4$    & \multicolumn{1}{l}{1.25E-10(6.32E-10)} &       &       & \multicolumn{1}{l}{2.12E-8(7.52E-8)} &       &       & \multicolumn{1}{l}{4.47E-22(1.66E-21)} &       &       & \multicolumn{1}{l}{4.73E-27(6.85E-28)} &       &       & \multicolumn{1}{l}{2.89E-3(1.28E-3)} &  \\
& $T_5$    & \multicolumn{1}{l}{9.95E+1(1.08E+2)} &       &       & \multicolumn{1}{l}{1.10E+2(1.68E+2)} &       &       & \multicolumn{1}{l}{6.33E+1(3.51E+1)} &       &       & \multicolumn{1}{l}{9.99E+1(1.35E+2)} &       &       & \multicolumn{1}{l}{3.26E+2(5.59E+2)} &  \\
\midrule
\multirow{5}[2]{*}{5} & $T_1$    & \multicolumn{1}{l}{2.56E+2(7.90E+1)} & \multicolumn{1}{c}{\multirow{5}[2]{*}{\textbf{ -6.60E+1 }}} &       & \multicolumn{1}{l}{3.54E+2(1.10E+2)} & \multicolumn{1}{c}{\multirow{5}[2]{*}{ -4.72E+1 }} &       & \multicolumn{1}{l}{3.33E+2(1.00E+2)} & \multicolumn{1}{c}{\multirow{5}[2]{*}{-8.87E+0}} &       & \multicolumn{1}{l}{2.50E+2(7.14E+1)} & \multicolumn{1}{c}{\multirow{5}[2]{*}{-4.12E+1}} &       & \multicolumn{1}{l}{6.23E+2(1.21E+2)} & \multicolumn{1}{c}{\multirow{5}[2]{*}{1.63E+2}} \\
& $T_2$    & \multicolumn{1}{l}{5.25E-3(8.46E-3)} &       &       & \multicolumn{1}{l}{6.24E-3(7.61E-3)} &       &       & \multicolumn{1}{l}{6.24E-3(8.15E-3)} &       &       & \multicolumn{1}{l}{5.34E-3(7.78E-3)} &       &       & \multicolumn{1}{l}{1.11E-2(1.11E-2)} &  \\
& $T_3$    & \multicolumn{1}{l}{2.50E+1(3.89E+0)} &       &       & \multicolumn{1}{l}{2.56E+1(3.16E+0)} &       &       & \multicolumn{1}{l}{3.22E+1(5.12E+0)} &       &       & \multicolumn{1}{l}{3.07E+1(4.10E+0)} &       &       & \multicolumn{1}{l}{4.25E+1(6.84E+0)} &  \\
& $T_4$    & \multicolumn{1}{l}{2.54E+2(9.16E+1)} &       &       & \multicolumn{1}{l}{3.08E+2(1.15E+2)} &       &       & \multicolumn{1}{l}{3.30E+2(9.99E+1)} &       &       & \multicolumn{1}{l}{2.46E+2(7.12E+1)} &       &       & \multicolumn{1}{l}{5.81E+2(9.87E+1)} &  \\
& $T_5$    & \multicolumn{1}{l}{6.65E-3(9.03E-3)} &       &       & \multicolumn{1}{l}{2.30E-3(4.96E-3)} &       &       & \multicolumn{1}{l}{5.99E-3(7.39E-3)} &       &       & \multicolumn{1}{l}{8.21E-3(9.01E-3)} &       &       & \multicolumn{1}{l}{1.09E-2(1.05E-2)} &  \\
\midrule
\multirow{5}[2]{*}{6} & $T_1$    & \multicolumn{1}{l}{9.72E+1(1.33E+2)} & \multicolumn{1}{c}{\multirow{5}[2]{*}{ -1.91E+1 }} &       & \multicolumn{1}{l}{2.01E+2(2.62E+2)} & \multicolumn{1}{c}{\multirow{5}[2]{*}{ 2.06E+1 }} &       & \multicolumn{1}{l}{2.20E+2(4.15E+2)} & \multicolumn{1}{c}{\multirow{5}[2]{*}{-7.12E+0}} &       & \multicolumn{1}{l}{1.07E+2(1.31E+2)} & \multicolumn{1}{c}{\multirow{5}[2]{*}{\textbf{-3.51E+1}}} &       & \multicolumn{1}{l}{8.20E+2(1.77E+3)} & \multicolumn{1}{c}{\multirow{5}[2]{*}{4.07E+1}} \\
& $T_2$    & \multicolumn{1}{l}{5.99E-3(8.14E-3)} &       &       & \multicolumn{1}{l}{8.13E-3(7.79E-3)} &       &       & \multicolumn{1}{l}{7.47E-3(1.04E-2)} &       &       & \multicolumn{1}{l}{6.32E-3(8.99E-3)} &       &       & \multicolumn{1}{l}{9.01E-3(8.89E-3)} &  \\
& $T_3$    & \multicolumn{1}{l}{9.51E+3(1.79E+3)} &       &       & \multicolumn{1}{l}{1.04E+4(1.62E+3)} &       &       & \multicolumn{1}{l}{9.31E+3(1.90E+3)} &       &       & \multicolumn{1}{l}{7.64E+3(1.49E+3)} &       &       & \multicolumn{1}{l}{8.01E+3(9.12E+2)} &  \\
& $T_4$    & \multicolumn{1}{l}{1.29E+2(1.68E+2)} &       &       & \multicolumn{1}{l}{1.28E+2(1.46E+2)} &       &       & \multicolumn{1}{l}{2.07E+2(3.36E+2)} &       &       & \multicolumn{1}{l}{5.23E+1(1.96E+1)} &       &       & \multicolumn{1}{l}{1.64E+3(2.60E+3)} &  \\
& $T_5$    & \multicolumn{1}{l}{4.43E-3(6.51E-3)} &       &       & \multicolumn{1}{l}{8.94E-3(1.22E-2)} &       &       & \multicolumn{1}{l}{5.67E-3(6.79E-3)} &       &       & \multicolumn{1}{l}{8.86E-3(9.17E-3)} &       &       & \multicolumn{1}{l}{8.67E-3(9.72E-3)} &  \\
\midrule
\multirow{5}[2]{*}{7} & $T_1$    & \multicolumn{1}{l}{3.66E+0(9.11E-1)} & \multicolumn{1}{c}{\multirow{5}[2]{*}{\textbf{ -6.67E+1 }}} &       & \multicolumn{1}{l}{3.34E+0(7.66E-1)} & \multicolumn{1}{c}{\multirow{5}[2]{*}{ -6.21E+1 }} &       & \multicolumn{1}{l}{3.16E+0(1.20E+0)} & \multicolumn{1}{c}{\multirow{5}[2]{*}{-2.98E+1}} &       & \multicolumn{1}{l}{4.36E+0(8.68E-1)} & \multicolumn{1}{c}{\multirow{5}[2]{*}{-2.95E+1}} &       & \multicolumn{1}{l}{5.39E+0(2.60E+0)} & \multicolumn{1}{c}{\multirow{5}[2]{*}{1.88E+2}} \\
& $T_2$    & \multicolumn{1}{l}{2.46E+2(6.87E+1)} &       &       & \multicolumn{1}{l}{3.58E+2(1.19E+2)} &       &       & \multicolumn{1}{l}{3.26E+2(1.12E+2)} &       &       & \multicolumn{1}{l}{2.51E+2(5.32E+1)} &       &       & \multicolumn{1}{l}{6.54E+2(1.51E+2)} &  \\
& $T_3$    & \multicolumn{1}{l}{2.64E+1(2.71E+0)} &       &       & \multicolumn{1}{l}{2.59E+1(3.26E+0)} &       &       & \multicolumn{1}{l}{3.24E+1(4.52E+0)} &       &       & \multicolumn{1}{l}{3.12E+1(4.54E+0)} &       &       & \multicolumn{1}{l}{4.47E+1(6.17E+0)} &  \\
& $T_4$    & \multicolumn{1}{l}{3.98E+0(8.81E-1)} &       &       & \multicolumn{1}{l}{3.42E+0(7.77E-1)} &       &       & \multicolumn{1}{l}{3.72E+0(1.03E+0)} &       &       & \multicolumn{1}{l}{4.46E+0(9.77E-1)} &       &       & \multicolumn{1}{l}{5.45E+0(2.68E+0)} &  \\
& $T_5$    & \multicolumn{1}{l}{2.62E+2(7.80E+1)} &       &       & \multicolumn{1}{l}{2.84E+2(1.02E+2)} &       &       & \multicolumn{1}{l}{3.46E+2(1.29E+2)} &       &       & \multicolumn{1}{l}{2.44E+2(5.00E+1)} &       &       & \multicolumn{1}{l}{5.96E+2(1.29E+2)} &  \\
\midrule
\multirow{5}[2]{*}{8} & $T_1$    & \multicolumn{1}{l}{1.97E+2(3.05E+2)} & \multicolumn{1}{c}{\multirow{5}[2]{*}{\textbf{ -4.79E+1 }}} &       & \multicolumn{1}{l}{3.92E+2(8.54E+2)} & \multicolumn{1}{c}{\multirow{5}[2]{*}{ -3.69E+1 }} &       & \multicolumn{1}{l}{3.48E+2(5.06E+2)} & \multicolumn{1}{c}{\multirow{5}[2]{*}{-4.60E+1}} &       & \multicolumn{1}{l}{1.94E+2(2.27E+2)} & \multicolumn{1}{c}{\multirow{5}[2]{*}{-4.78E+1}} &       & \multicolumn{1}{l}{1.10E+3(2.23E+3)} & \multicolumn{1}{c}{\multirow{5}[2]{*}{1.79E+2}} \\
& $T_2$    & \multicolumn{1}{l}{5.25E+0(1.13E+0)} &       &       & \multicolumn{1}{l}{4.39E+0(1.24E+0)} &       &       & \multicolumn{1}{l}{3.67E+0(1.37E+0)} &       &       & \multicolumn{1}{l}{4.47E+0(7.68E-1)} &       &       & \multicolumn{1}{l}{2.00E+1(8.09E-2)} &  \\
& $T_3$    & \multicolumn{1}{l}{3.77E+2(1.20E+2)} &       &       & \multicolumn{1}{l}{4.29E+2(1.78E+2)} &       &       & \multicolumn{1}{l}{3.40E+2(7.08E+1)} &       &       & \multicolumn{1}{l}{2.56E+2(7.38E+1)} &       &       & \multicolumn{1}{l}{6.35E+2(1.31E+2)} &  \\
& $T_4$    & \multicolumn{1}{l}{5.91E-3(6.49E-3)} &       &       & \multicolumn{1}{l}{6.81E-3(9.70E-3)} &       &       & \multicolumn{1}{l}{5.42E-3(7.52E-3)} &       &       & \multicolumn{1}{l}{5.99E-3(7.10E-3)} &       &       & \multicolumn{1}{l}{1.11E-2(8.90E-3)} &  \\
& $T_5$    & \multicolumn{1}{l}{3.05E+1(3.74E+0)} &       &       & \multicolumn{1}{l}{2.97E+1(3.22E+0)} &       &       & \multicolumn{1}{l}{3.51E+1(5.04E+0)} &       &       & \multicolumn{1}{l}{3.87E+1(6.72E+0)} &       &       & \multicolumn{1}{l}{5.40E+1(5.22E+0)} &  \\
\midrule
\multirow{5}[1]{*}{9} & $T_1$    & \multicolumn{1}{l}{5.47E+0(1.70E+0)} & \multicolumn{1}{c}{\multirow{5}[1]{*}{ -1.99E+1 }} &       & \multicolumn{1}{l}{4.95E+0(1.29E+0)} & \multicolumn{1}{c}{\multirow{5}[1]{*}{ -6.90E+0 }} &       & \multicolumn{1}{l}{3.86E+0(1.26E+0)} & \multicolumn{1}{c}{\multirow{5}[1]{*}{-2.67E+1}} &       & \multicolumn{1}{l}{4.35E+0(1.01E+0)} & \multicolumn{1}{c}{\multirow{5}[1]{*}{\textbf{-7.26E+1}}} &       & \multicolumn{1}{l}{2.00E+1(7.78E-2)} & \multicolumn{1}{c}{\multirow{5}[1]{*}{1.26E+2}} \\
& $T_2$    & \multicolumn{1}{l}{3.65E+2(1.10E+2)} &       &       & \multicolumn{1}{l}{4.42E+2(1.28E+2)} &       &       & \multicolumn{1}{l}{3.95E+2(1.19E+2)} &       &       & \multicolumn{1}{l}{2.54E+2(8.27E+1)} &       &       & \multicolumn{1}{l}{6.50E+2(1.45E+2)} &  \\
& $T_3$    & \multicolumn{1}{l}{1.23E-2(1.83E-2)} &       &       & \multicolumn{1}{l}{7.80E-3(9.10E-3)} &       &       & \multicolumn{1}{l}{9.44E-3(8.69E-3)} &       &       & \multicolumn{1}{l}{6.07E-3(8.46E-3)} &       &       & \multicolumn{1}{l}{1.04E-2(9.54E-3)} &  \\
& $T_4$    & \multicolumn{1}{l}{3.38E+1(3.80E+0)} &       &       & \multicolumn{1}{l}{2.98E+1(3.42E+0)} &       &       & \multicolumn{1}{l}{3.50E+1(5.30E+0)} &       &       & \multicolumn{1}{l}{3.94E+1(5.99E+0)} &       &       & \multicolumn{1}{l}{5.33E+1(5.41E+0)} &  \\
& $T_5$    & \multicolumn{1}{l}{8.15E+3(1.43E+3)} &       &       & \multicolumn{1}{l}{9.80E+3(1.28E+3)} &       &       & \multicolumn{1}{l}{8.11E+3(1.86E+3)} &       &       & \multicolumn{1}{l}{6.30E+3(1.06E+3)} &       &       & \multicolumn{1}{l}{6.36E+3(8.44E+2)} &  \\
\midrule
\multicolumn{2}{c}{Mean} & -     & \textbf{-5.26E+1} &       & -     & -4.07E+1 &       & -     & -2.53E+1 &       & -     & -4.74E+1 &       & -     & 1.66E+2 \\

\bottomrule
	\end{tabular}%
}
	\label{tab:cmp_many_5Tasks}%
\end{table*}%

In Table \ref{tab:cmp_many_5Tasks}, the SaMTPSO-S1 achieves the best mean score $-5.26E+1$ comparing to the other algorithms. On problem 1, 3, 5, 7 and 8, the SaMTPSO-S1 achieves the best scores. Comparing to the SaMTPSO-S1 and the SREMTO, the performance of the SaMTPSO-S2 is slightly inferior, achieving a mean score $-4.07E+1$. For the SREMTO, it achieves a similarity on problem 1 comparing to that of the SaMTPSO-S1 and the SaMTPSO-S2, and achieves the best results on problem 6 and 9, on which the SaMTPSO-S1 and the SaMTPSO-S2 have achieved not bad results. Therefore, the proposed SaMTPSO is still promising in handling these kinds of MTO problems.

\subsubsection{Parameters Analysis}

This experiment studies the two key parameters in the proposed SaMTPSO, i.e., $bp$ and $LP$. $bp$ is a parameter to assign a small probability to the knowledge sources of a task's pool, so that all sources can be chosen in each generation. $LP$ is the learning period in the proposed SaMTPSO as previously described. The experiment is conducted by comparing the experimental results of the SaMTPSO-S1 with different parameter settings on these two parameters. The 9 popular MTO test problems in test suite 1 is employed. In Table \ref{tab:benchmarkset:params_bp} and \ref{tab:benchmarkset:params_LP}, the means, bracketed standard deviations of the FEVs over 30 runs are reported along with the achieved scores by the SaMTPSO-S1 of different parameter settings. The analysis results regarding to SaMTPSO-S2 is similar to that of the SaMTPSO-S1. Therefore, this experiment only demonstrates the analysis on SaMTPSO-S1.

\begin{table*}[htb!]
	\centering
	\caption{Achieved FEVs (mean and bracketed standard deviations) and scores by the SaMTPSO-S1 using different $bp$ settings.}
	\resizebox{516pt}{90pt}{ %
		\begin{tabular}{C{0.04\textwidth} C{0.03\textwidth} C{0.12\textwidth} C{0.06\textwidth}C{0.000001\textwidth} C{0.12\textwidth}C{0.06\textwidth} C{0.000001\textwidth} C{0.12\textwidth} C{0.06\textwidth}C{0.000001\textwidth}C{0.12\textwidth} C{0.06\textwidth}C{0.000001\textwidth}C{0.12\textwidth} C{0.06\textwidth}C{0.000001\textwidth} C{0.12\textwidth}C{0.06\textwidth}C{0.000001\textwidth} C{0.12\textwidth}C{0.06\textwidth}}
			\toprule
			
			\multirow{2}[4]{*}{ \tabincell{c}{Problem \\ Sets\\}} & \multirow{2}[4]{*}{Task} & \multicolumn{2}{c}{bp=0.0001} &       & \multicolumn{2}{c}{bp=0.0005} &       & \multicolumn{2}{c}{bp=0.001} &       & \multicolumn{2}{c}{bp=0.005} &       & \multicolumn{2}{c}{bp=0.01} &       & \multicolumn{2}{c}{bp=0.05} &       & \multicolumn{2}{c}{bp=0.1} \\
			\cmidrule{3-4}\cmidrule{6-7}\cmidrule{9-10}\cmidrule{12-13}\cmidrule{15-16}\cmidrule{18-19}\cmidrule{21-22}          &       & Mean(Std) & Score &       & Mean(Std) & Score &       & Mean(Std) & Score &       & Mean(Std) & Score &       & Mean(Std) & Score &       & Mean(Std) & Score &       & Mean(Std) & Score \\
			\midrule

    \multirow{2}[2]{*}{1} & $T_1$    & 3.30E-2(5.10E-2) & \multirow{2}[2]{*}{-2.97E+1} &       & 8.10E-3(8.85E-3) & \multirow{2}[2]{*}{-3.49E+1} &       & 6.00E-3(7.70E-3) & \multirow{2}[2]{*}{-3.44E+1} &       & 5.39E-3(8.21E-3) & \multirow{2}[2]{*}{\textbf{-3.88E+1}} &       & 1.20E-2(2.79E-2) & \multirow{2}[2]{*}{-3.27E+1} &       & 3.92E-1(1.79E-1) & \multirow{2}[2]{*}{8.21E+1} &       & 4.34E-1(1.37E-1) & \multirow{2}[2]{*}{8.84E+1} \\
& $T_2$    & 2.76E+1(2.29E+1) &       &       & 2.41E+1(1.84E+1) &       &       & 2.59E+1(2.20E+1) &       &       & 1.54E+1(1.65E+1) &       &       & 2.80E+1(4.44E+1) &       &       & 1.70E+2(4.12E+1) &       &       & 1.70E+2(4.22E+1) &  \\
\midrule
\multirow{2}[2]{*}{2} & $T_1$    & 2.84E+0(9.60E-1) & \multirow{2}[2]{*}{-1.50E+1} &       & 2.89E+0(5.63E-1) & \multirow{2}[2]{*}{-1.43E+1} &       & 2.79E+0(6.59E-1) & \multirow{2}[2]{*}{-1.99E+1} &       & 2.65E+0(6.42E-1) & \multirow{2}[2]{*}{\textbf{-3.00E+1}} &       & 2.79E+0(5.51E-1) & \multirow{2}[2]{*}{-1.95E+1} &       & 3.02E+0(5.79E-1) & \multirow{2}[2]{*}{4.81E+1} &       & 3.01E+0(4.98E-1) & \multirow{2}[2]{*}{5.05E+1} \\
& $T_2$    & 5.12E+1(3.10E+1) &       &       & 4.80E+1(1.54E+1) &       &       & 4.54E+1(1.72E+1) &       &       & 3.84E+1(1.61E+1) &       &       & 4.67E+1(1.74E+1) &       &       & 1.68E+2(4.35E+1) &       &       & 1.74E+2(4.62E+1) &  \\
\midrule
\multirow{2}[2]{*}{3} & $T_1$    & 7.81E-1(3.85E+0) & \multirow{2}[2]{*}{1.56E+1} &       & 8.27E-2(1.27E-1) & \multirow{2}[2]{*}{-6.90E+0} &       & 4.90E-2(6.78E-2) & \multirow{2}[2]{*}{-6.79E+0} &       & 1.49E+0(5.31E+0) & \multirow{2}[2]{*}{1.79E+1} &       & 8.12E-2(1.17E-1) & \multirow{2}[2]{*}{-6.86E+0} &       & 1.67E-1(2.92E-1) & \multirow{2}[2]{*}{-5.82E+0} &       & 6.18E-2(7.73E-2) & \multirow{2}[2]{*}{\textbf{-7.17E+0}} \\
& $T_2$    & 1.42E+2(7.32E+2) &       &       & 2.86E-1(7.88E-1) &       &       & 5.45E+0(2.26E+1) &       &       & 7.99E+1(3.03E+2) &       &       & 9.41E-1(3.40E+0) &       &       & 1.06E+0(2.47E+0) &       &       & 1.46E-1(2.31E-1) &  \\
\midrule
\multirow{2}[2]{*}{4} & $T_1$    & 3.69E+2(8.71E+1) & \multirow{2}[2]{*}{-3.32E+1} &       & 3.50E+2(1.05E+2) & \multirow{2}[2]{*}{-3.66E+1} &       & 3.66E+2(1.37E+2) & \multirow{2}[2]{*}{-3.38E+1} &       & 3.47E+2(9.04E+1) & \multirow{2}[2]{*}{\textbf{-3.70E+1}} &       & 4.67E+2(1.43E+2) & \multirow{2}[2]{*}{-1.63E+1} &       & 6.78E+2(8.10E+1) & \multirow{2}[2]{*}{7.31E+1} &       & 6.81E+2(7.85E+1) & \multirow{2}[2]{*}{8.38E+1} \\
& $T_2$    & 3.19E-2(7.47E-2) &       &       & 7.01E-4(1.76E-3) &       &       & 4.37E-5(1.45E-4) &       &       & 2.91E-3(1.25E-2) &       &       & 3.05E+0(8.10E+0) &       &       & 6.09E+2(3.35E+2) &       &       & 7.24E+2(2.55E+2) &  \\
\midrule
\multirow{2}[2]{*}{5} & $T_1$    & 1.11E+0(9.34E-1) & \multirow{2}[2]{*}{\textbf{-6.12E-1}} &       & 9.01E-1(8.61E-1) & \multirow{2}[2]{*}{-1.72E+1} &       & 1.06E+0(9.51E-1) & \multirow{2}[2]{*}{-1.09E+1} &       & 9.78E-1(8.56E-1) & \multirow{2}[2]{*}{-1.80E+1} &       & 9.16E-1(8.39E-1) & \multirow{2}[2]{*}{-1.69E+1} &       & 1.02E+0(6.89E-1) & \multirow{2}[2]{*}{1.75E+1} &       & 1.28E+0(7.92E-1) & \multirow{2}[2]{*}{4.61E+1} \\
& $T_2$    & 1.15E+2(3.66E+1) &       &       & 9.64E+1(2.72E+1) &       &       & 9.78E+1(2.85E+1) &       &       & 8.90E+1(2.14E+1) &       &       & 9.58E+1(1.92E+1) &       &       & 1.59E+2(6.64E+1) &       &       & 1.99E+2(9.28E+1) &  \\
\midrule
\multirow{2}[2]{*}{6} & $T_1$    & 4.67E+0(1.46E+0) & \multirow{2}[2]{*}{-1.18E+1} &       & 4.26E+0(1.38E+0) & \multirow{2}[2]{*}{\textbf{-4.00E+1}} &       & 4.14E+0(9.21E-1) & \multirow{2}[2]{*}{-3.83E+1} &       & 5.40E+0(1.28E+0) & \multirow{2}[2]{*}{-9.06E+0} &       & 5.75E+0(1.67E+0) & \multirow{2}[2]{*}{3.01E+0} &       & 8.06E+0(1.47E+0) & \multirow{2}[2]{*}{4.52E+1} &       & 8.13E+0(1.48E+0) & \multirow{2}[2]{*}{5.09E+1} \\
& $T_2$    & 4.38E+0(3.13E+0) &       &       & 2.89E+0(1.98E+0) &       &       & 3.12E+0(1.59E+0) &       &       & 3.86E+0(1.41E+0) &       &       & 4.33E+0(1.18E+0) &       &       & 4.94E+0(1.46E+0) &       &       & 5.25E+0(1.59E+0) &  \\
\midrule
\multirow{2}[2]{*}{7} & $T_1$    & 1.35E+2(5.08E+1) & \multirow{2}[2]{*}{-2.98E+1} &       & 1.34E+2(5.85E+1) & \multirow{2}[2]{*}{-2.77E+1} &       & 1.18E+2(4.29E+1) & \multirow{2}[2]{*}{-3.21E+1} &       & 8.56E+1(3.93E+1) & \multirow{2}[2]{*}{\textbf{-3.51E+1}} &       & 9.75E+1(6.25E+1) & \multirow{2}[2]{*}{-3.08E+1} &       & 6.31E+2(3.40E+2) & \multirow{2}[2]{*}{6.78E+1} &       & 7.84E+2(4.97E+2) & \multirow{2}[2]{*}{8.77E+1} \\
& $T_2$    & 3.58E+1(1.45E+1) &       &       & 4.11E+1(2.18E+1) &       &       & 3.38E+1(1.22E+1) &       &       & 3.31E+1(1.78E+1) &       &       & 4.09E+1(4.18E+1) &       &       & 1.69E+2(5.68E+1) &       &       & 1.86E+2(4.05E+1) &  \\
\midrule
\multirow{2}[2]{*}{8} & $T_1$    & 4.81E-2(6.88E-2) & \multirow{2}[2]{*}{-1.34E+1} &       & 1.10E-2(1.01E-2) & \multirow{2}[2]{*}{-2.90E+1} &       & 7.92E-3(9.54E-3) & \multirow{2}[2]{*}{\textbf{-3.80E+1}} &       & 7.36E-3(1.00E-2) & \multirow{2}[2]{*}{-2.47E+1} &       & 8.94E-2(1.13E-1) & \multirow{2}[2]{*}{-1.28E+1} &       & 1.08E+0(5.13E-2) & \multirow{2}[2]{*}{5.60E+1} &       & 1.08E+0(4.71E-2) & \multirow{2}[2]{*}{6.18E+1} \\
& $T_2$    & 1.95E+1(4.09E+0) &       &       & 1.84E+1(2.16E+0) &       &       & 1.76E+1(2.35E+0) &       &       & 1.88E+1(2.32E+0) &       &       & 1.93E+1(1.66E+0) &       &       & 1.99E+1(1.65E+0) &       &       & 2.04E+1(1.92E+0) &  \\
\midrule
\multirow{2}[2]{*}{9} & $T_1$    & 3.63E+2(1.11E+2) & \multirow{2}[2]{*}{-1.63E+1} &       & 3.98E+2(1.24E+2) & \multirow{2}[2]{*}{-2.27E+1} &       & 4.04E+2(1.18E+2) & \multirow{2}[2]{*}{\textbf{-7.96E+0}} &       & 4.70E+2(1.70E+2) & \multirow{2}[2]{*}{-1.70E+1} &       & 5.47E+2(2.14E+2) & \multirow{2}[2]{*}{-1.86E+1} &       & 1.78E+3(3.47E+2) & \multirow{2}[2]{*}{4.07E+1} &       & 1.83E+3(2.31E+2) & \multirow{2}[2]{*}{4.18E+1} \\
& $T_2$    & 2.09E+2(4.56E+2) &       &       & 8.38E+1(1.85E+2) &       &       & 3.09E+2(1.06E+3) &       &       & 1.21E+2(3.14E+2) &       &       & 4.16E+1(1.18E+2) &       &       & 8.29E+1(1.51E+2) &       &       & 6.53E+1(1.63E+2) &  \\
\midrule
\multicolumn{2}{c}{Mean} & -     & -1.49E+1 &       & -     & \textbf{-2.55E+1} &       & -     & -2.47E+1 &       & -     & -2.13E+1 &       & -     & -1.68E+1 &       & -     & 4.72E+1 &       & -     & 5.60E+1 \\

			\bottomrule
		\end{tabular}%
	}
	\label{tab:benchmarkset:params_bp}%
\end{table*}%

Table \ref{tab:benchmarkset:params_bp} shows that the mean scores of the SaMTPSO-S1 are getting worse as the $bp$ settings increasing from $0.0001$ to $0.1$. When $bp=0.0005$, the SaMTPSO-S1 achieves the best results, achieving a mean score $-2.55E+1$. $bp$ is a parameter to assign a small probability to the knowledge sources of a task's pool. The bigger the $bp$ settings, the smaller the self-adaptive ability of the SaMTPSO. And all knowledge sources of a task's pool will tend to be chosen with equal probability. Though the SaMTPSO-S1 achieves best results when $bp=0.0005$, we think it may be too small for the SaMTPSO. Therefore, $bp=0.001$ is chosen as a most suitable setting for the SaMTPSO, because the achieved results is not bad as well.

\begin{table*}[htb!]
	\centering
	\caption{Achieved FEVs (mean and bracketed standard deviations) and scores by the SaMTPSO-S1 using different $LP$ settings.}
	\resizebox{516pt}{90pt}{ %
		\begin{tabular}{C{0.04\textwidth} C{0.03\textwidth} C{0.12\textwidth} C{0.05\textwidth}C{0.000001\textwidth} C{0.12\textwidth}C{0.06\textwidth} C{0.000001\textwidth} C{0.12\textwidth} C{0.06\textwidth}C{0.000001\textwidth}C{0.12\textwidth} C{0.06\textwidth}C{0.000001\textwidth}C{0.12\textwidth} C{0.06\textwidth}C{0.000001\textwidth} C{0.12\textwidth}C{0.06\textwidth}C{0.000001\textwidth} C{0.12\textwidth}C{0.06\textwidth}}
			\toprule
			
			\multirow{3}{*}{ \tabincell{c}{Problem \\ Sets\\}} & \multirow{2}[4]{*}{Task} & \multicolumn{2}{c}{LP=1} &       & \multicolumn{2}{c}{LP=2} &       & \multicolumn{2}{c}{LP=5} &       & \multicolumn{2}{c}{LP=10} &       & \multicolumn{2}{c}{LP=20} &       & \multicolumn{2}{c}{LP=50} &       & \multicolumn{2}{c}{LP=100} \\
			\cmidrule{3-4}\cmidrule{6-7}\cmidrule{9-10}\cmidrule{12-13}\cmidrule{15-16}\cmidrule{18-19}\cmidrule{21-22}          &       & Mean(Std) & Score &       & Mean(Std) & Score &       & Mean(Std) & Score &       & Mean(Std) & Score &       & Mean(Std) & Score &       & Mean(Std) & Score &       & Mean(Std) & Score \\
			\midrule
			
    \multirow{2}[2]{*}{1} & $T1$    & 5.88E-2(6.69E-2) & \multirow{2}[2]{*}{8.50E+1} &       & 6.16E-3(6.49E-3) & \multirow{2}[2]{*}{-6.74E+0} &       & 6.39E-3(8.13E-3) & \multirow{2}[2]{*}{-1.52E+1} &       & 6.00E-3(7.70E-3) & \multirow{2}[2]{*}{\textbf{-1.96E+1}} &       & 1.10E-2(1.03E-2) & \multirow{2}[2]{*}{-1.81E+1} &       & 1.16E-2(1.36E-2) & \multirow{2}[2]{*}{-6.88E+0} &       & 1.05E-2(9.92E-3) & \multirow{2}[2]{*}{-1.84E+1} \\
& $T_2$    & 1.23E+2(8.92E+1) &       &       & 4.86E+1(1.98E+1) &       &       & 3.31E+1(1.82E+1) &       &       & 2.59E+1(2.20E+1) &       &       & 2.02E+1(1.75E+1) &       &       & 3.93E+1(5.06E+1) &       &       & 2.06E+1(1.79E+1) &  \\
\midrule
\multirow{2}[2]{*}{2} & $T1$    & 4.72E+0(1.46E+0) & \multirow{2}[2]{*}{1.16E+2} &       & 3.07E+0(7.11E-1) & \multirow{2}[2]{*}{-2.21E+0} &       & 2.98E+0(5.30E-1) & \multirow{2}[2]{*}{-1.26E+1} &       & 2.79E+0(6.59E-1) & \multirow{2}[2]{*}{-2.02E+1} &       & 2.78E+0(4.73E-1) & \multirow{2}[2]{*}{-2.06E+1} &       & 2.55E+0(4.65E-1) & \multirow{2}[2]{*}{\textbf{-3.01E+1}} &       & 2.52E+0(6.38E-1) & \multirow{2}[2]{*}{-3.00E+1} \\
& $T_2$    & 2.51E+2(5.76E+1) &       &       & 7.07E+1(2.91E+1) &       &       & 5.12E+1(1.70E+1) &       &       & 4.54E+1(1.72E+1) &       &       & 4.55E+1(1.20E+1) &       &       & 3.80E+1(1.39E+1) &       &       & 4.08E+1(1.99E+1) &  \\
\midrule
\multirow{2}[2]{*}{3} & $T1$    & 7.86E+0(1.03E+1) & \multirow{2}[2]{*}{5.33E+1} &       & 8.05E-1(3.84E+0) & \multirow{2}[2]{*}{2.82E+0} &       & 6.15E-2(9.46E-2) & \multirow{2}[2]{*}{-8.25E+0} &       & 4.90E-2(6.78E-2) & \multirow{2}[2]{*}{-1.10E+1} &       & 7.74E-2(8.33E-2) & \multirow{2}[2]{*}{\textbf{-1.24E+1}} &       & 1.11E-1(1.85E-1) & \multirow{2}[2]{*}{-1.21E+1} &       & 8.00E-2(1.38E-1) & \multirow{2}[2]{*}{-1.23E+1} \\
& $T_2$    & 6.03E+1(2.33E+2) &       &       & 3.61E+1(1.09E+2) &       &       & 1.44E+1(4.46E+1) &       &       & 5.45E+0(2.26E+1) &       &       & 2.47E-1(3.89E-1) &       &       & 5.56E-1(1.15E+0) &       &       & 3.54E-1(7.90E-1) &  \\
\midrule
\multirow{2}[2]{*}{4} & $T1$    & 3.94E+2(1.28E+2) & \multirow{2}[2]{*}{7.41E+1} &       & 3.64E+2(9.16E+1) & \multirow{2}[2]{*}{-1.35E+1} &       & 3.71E+2(1.24E+2) & \multirow{2}[2]{*}{-1.18E+1} &       & 3.66E+2(1.37E+2) & \multirow{2}[2]{*}{-1.30E+1} &       & 3.56E+2(8.17E+1) & \multirow{2}[2]{*}{\textbf{-1.58E+1}} &       & 3.93E+2(9.34E+1) & \multirow{2}[2]{*}{-5.51E+0} &       & 3.61E+2(9.84E+1) & \multirow{2}[2]{*}{-1.45E+1} \\
& $T_2$    & 8.30E+2(3.11E+2) &       &       & 1.87E-6(5.79E-6) &       &       & 4.46E-5(1.22E-4) &       &       & 4.37E-5(1.45E-4) &       &       & 2.01E-4(6.22E-4) &       &       & 1.36E-2(5.75E-2) &       &       & 5.68E-3(1.55E-2) &  \\
\midrule
\multirow{2}[2]{*}{5} & $T1$    & 4.28E+0(1.29E+0) & \multirow{2}[2]{*}{1.07E+2} &       & 1.50E+0(7.70E-1) & \multirow{2}[2]{*}{-1.03E+1} &       & 1.12E+0(8.92E-1) & \multirow{2}[2]{*}{-1.74E+1} &       & 1.06E+0(9.51E-1) & \multirow{2}[2]{*}{-1.89E+1} &       & 9.86E-1(8.93E-1) & \multirow{2}[2]{*}{-2.03E+1} &       & 1.14E+0(9.13E-1) & \multirow{2}[2]{*}{-1.70E+1} &       & 8.35E-1(8.56E-1) & \multirow{2}[2]{*}{\textbf{-2.35E+1}} \\
& $T_2$    & 1.31E+3(1.14E+3) &       &       & 9.18E+1(2.68E+1) &       &       & 1.01E+2(2.42E+1) &       &       & 9.78E+1(2.85E+1) &       &       & 9.90E+1(2.87E+1) &       &       & 1.04E+2(3.06E+1) &       &       & 9.79E+1(2.51E+1) &  \\
\midrule
\multirow{2}[2]{*}{6} & $T1$    & 4.34E+0(1.25E+0) & \multirow{2}[2]{*}{\textbf{-2.83E+1}} &       & 4.13E+0(7.29E-1) & \multirow{2}[2]{*}{-1.22E+1} &       & 4.11E+0(7.76E-1) & \multirow{2}[2]{*}{-1.99E+1} &       & 4.14E+0(9.21E-1) & \multirow{2}[2]{*}{-1.91E+1} &       & 5.06E+0(1.21E+0) & \multirow{2}[2]{*}{-6.68E+0} &       & 6.53E+0(2.10E+0) & \multirow{2}[2]{*}{4.84E+1} &       & 6.06E+0(1.66E+0) & \multirow{2}[2]{*}{3.77E+1} \\
& $T_2$    & 2.45E+0(6.87E-1) &       &       & 3.49E+0(1.42E+0) &       &       & 3.11E+0(1.74E+0) &       &       & 3.12E+0(1.59E+0) &       &       & 2.87E+0(1.29E+0) &       &       & 4.29E+0(1.54E+0) &       &       & 4.19E+0(1.47E+0) &  \\
\midrule
\multirow{2}[2]{*}{7} & $T1$    & 1.94E+3(8.29E+3) & \multirow{2}[2]{*}{8.30E+1} &       & 1.04E+2(4.01E+1) & \multirow{2}[2]{*}{-6.77E+0} &       & 1.03E+2(4.13E+1) & \multirow{2}[2]{*}{-1.33E+1} &       & 1.18E+2(4.29E+1) & \multirow{2}[2]{*}{\textbf{-1.67E+1}} &       & 1.21E+2(5.53E+1) & \multirow{2}[2]{*}{-1.55E+1} &       & 1.26E+2(4.58E+1) & \multirow{2}[2]{*}{-1.63E+1} &       & 1.40E+2(5.10E+1) & \multirow{2}[2]{*}{-1.45E+1} \\
& $T_2$    & 2.16E+2(4.80E+1) &       &       & 5.60E+1(2.11E+1) &       &       & 4.16E+1(1.71E+1) &       &       & 3.38E+1(1.22E+1) &       &       & 3.64E+1(2.24E+1) &       &       & 3.45E+1(1.25E+1) &       &       & 3.81E+1(1.25E+1) &  \\
\midrule
\multirow{2}[2]{*}{8} & $T1$    & 8.23E-2(1.12E-1) & \multirow{2}[2]{*}{4.39E+1} &       & 7.21E-3(8.21E-3) & \multirow{2}[2]{*}{-4.65E+0} &       & 7.19E-3(8.50E-3) & \multirow{2}[2]{*}{-1.70E+1} &       & 7.92E-3(9.54E-3) & \multirow{2}[2]{*}{\textbf{-2.59E+1}} &       & 8.70E-3(1.19E-2) & \multirow{2}[2]{*}{-9.95E+0} &       & 2.96E-2(6.60E-2) & \multirow{2}[2]{*}{4.57E+0} &       & 3.87E-2(4.21E-2) & \multirow{2}[2]{*}{9.00E+0} \\
& $T_2$    & 2.00E+1(1.26E+0) &       &       & 1.93E+1(2.16E+0) &       &       & 1.83E+1(1.93E+0) &       &       & 1.76E+1(2.35E+0) &       &       & 1.88E+1(2.45E+0) &       &       & 1.91E+1(2.58E+0) &       &       & 1.90E+1(2.33E+0) &  \\
\midrule
\multirow{2}[2]{*}{9} & $T1$    & 3.63E+2(9.06E+1) & \multirow{2}[2]{*}{\textbf{-1.27E+1}} &       & 4.41E+2(1.22E+2) & \multirow{2}[2]{*}{5.74E+0} &       & 4.18E+2(1.26E+2) & \multirow{2}[2]{*}{2.10E+0} &       & 4.04E+2(1.18E+2) & \multirow{2}[2]{*}{6.89E+0} &       & 3.93E+2(9.36E+1) & \multirow{2}[2]{*}{-9.46E+0} &       & 3.93E+2(1.52E+2) & \multirow{2}[2]{*}{-1.16E+1} &       & 4.92E+2(1.40E+2) & \multirow{2}[2]{*}{1.90E+1} \\
& $T_2$    & 1.29E+2(3.06E+2) &       &       & 1.27E+2(2.17E+2) &       &       & 1.60E+2(4.73E+2) &       &       & 3.09E+2(1.06E+3) &       &       & 5.72E+1(1.84E+2) &       &       & 1.80E+1(5.07E+1) &       &       & 1.47E+2(7.72E+2) &  \\
\midrule
\multicolumn{2}{c}{Mean} & -     & 5.79E+1 &       & -     & -5.31E+0 &       & -     & -1.26E+1 &       & -     & \textbf{-1.53E+1} &       & -     & -1.43E+1 &       & -     & -5.16E+0 &       & -     & -5.27E+0 \\

			\bottomrule
		\end{tabular}%
	}
	\label{tab:benchmarkset:params_LP}%
\end{table*}%

Table \ref{tab:benchmarkset:params_LP} shows that, the mean scores of the SaMTPSO-S1 are getting worse as the $LP$ settings increasing from $2$ to $100$, and $LP=10$ is the best parameter setting for the algorithm. $LP=1$ means that the learned probabilities on $K$ knowledge sources for every component task may probably be impacted by randomness easily, because only the experience in the current generation is used. Therefore, $LP=10$ is chosen as a most suitable setting for the SaMTPSO. 

\section{Concluding Remarks}

This paper proposes a novel SaMTPSO, in which three novel strategies for adaptive knowledge transfer are developed, i.e., the knowledge transfer adaptation strategy, the focus search strategy and the knowledge incorporation strategy. Two versions of the SaMTPSO are developed, each employing a different form of knowledge incorporation strategy. To demonstrate the superiority of the SaMTPSO, several numerical experiments are conducted on two test suites, which compares the results of the SaMTPSO to that of several typical recently proposed EMTO algorithms. Besides, the last experiment has analyzed the influence of the settings of two key parameters in the SaMTPSO.

In the future, the framework of SaMTPSO can be introduced into many EAs to develop more powerful adaptive EMTO solvers. Further, more studies are needed to apply the SaMTPSO into solving real-world optimization problems, such as path planning problem\cite{bortoff2000path}, large-scale optimization problems\cite{tsurkov2013large}, multi-object optimization problems, etc.

\bibliographystyle{IEEEtran}
\bibliography{RefDatabase}

\end{document}